%% file: main.tex
\documentclass[sigconf]{acmart}

\AtBeginDocument{%
  }

\copyrightyear{2023}
\acmYear{2023}
\setcopyright{acmlicensed}
\acmConference[KDD '23]{Proceedings of the 29th ACM SIGKDD Conference on Knowledge Discovery and Data Mining}{August 6--10, 2023}{Long Beach, CA, USA}
\acmBooktitle{Proceedings of the 29th ACM SIGKDD Conference on Knowledge Discovery and Data Mining (KDD '23), August 6--10, 2023, Long Beach, CA, USA}
\acmPrice{15.00}
\acmDOI{10.1145/3580305.3599413} \acmISBN{979-8-4007-0103-0/23/08}

\input{dfn.tex}

\input{math_commands.tex}

\begin{document}

\title{Less is More: \method for Accurate, Robust, and Interpretable Graph Mining}

\settopmatter{authorsperrow=4}

\author{Jaemin Yoo}
\authornote{Both authors contributed equally to this research.}
\affiliation{%
  \institution{Carnegie Mellon University}
  \city{Pittsburgh}
  \country{USA}
}
\email{jaeminyoo@cmu.edu}

\author{Meng-Chieh Lee}
\authornotemark[1]
\affiliation{%
  \institution{Carnegie Mellon University}
  \city{Pittsburgh}
  \country{USA}
}
\email{mengchil@cs.cmu.edu}

\author{Shubhranshu Shekhar}
\affiliation{%
  \institution{Carnegie Mellon University}
  \city{Pittsburgh}
  \country{USA}
}
\email{shubhras@andrew.cmu.edu}

\author{Christos Faloutsos}
\affiliation{%
  \institution{Carnegie Mellon University}
  \city{Pittsburgh}
  \country{USA}
}
\email{christos@cs.cmu.edu}


\begin{abstract}
\input{000abstract.tex}
\end{abstract}




\maketitle

\input{010intro.tex}
\input{020related.tex}
\input{030linear.tex}
\input{040method.tex}
\input{050sanity.tex}
\input{060exp.tex}
\input{070conclusion.tex}

\begin{acks}
This work was partially funded by project AIDA (reference POCI-01-0247-FEDER-045907) under CMU Portugal, and a gift from PNC.
\end{acks}

\clearpage
\bibliographystyle{ACM-Reference-Format}
\bibliography{main}

\clearpage
\appendix
\input{110proof}
\input{120linear}
\input{130sanity}
\input{140exp}

\end{document}

%% file: dfn.tex
\usepackage{booktabs}
\usepackage{tablefootnote}
\usepackage{amsmath}
\usepackage{algorithm}
\usepackage{xspace}
\usepackage{subcaption}
\usepackage{multirow}
\usepackage{multicol}
\usepackage{paralist}

\usepackage{bbm}
\usepackage{tabularx, booktabs}
\usepackage{graphicx}
\newcolumntype{C}[1]{>{\centering\let\newline\\\arraybackslash\hspace{0pt}}m{#1}}
\newcolumntype{R}[1]{>{\raggedleft\let\newline\\\arraybackslash\hspace{0pt}}m{#1}}
\newcolumntype{L}[1]{>{\raggedright\let\newline\\\arraybackslash\hspace{0pt}}m{#1}}
\usepackage{colortbl}

\newcommand{\method}{{\textsc{SlimG}}\xspace}
\newcommand{\principle}{{\em KICS}\xspace}
\newcommand{\principleLong}{{\em careful simplicity}\xspace}
\renewcommand{\principle}{\principleLong}

\newcommand{\framework}{\textsc{GnnExp}\xspace}

\newcommand{\mycontribution}[1]{{\bf #1}\xspace}

\newcommand{\AsymK}[1]{\tilde{\mathbf{A}}_\mathrm{sym}^{#1}}

\hypersetup{
     colorlinks   = true,
     citecolor    = blue
}
\newcommand{\spaceBelowTableCaption}[0]{\vspace{-2.5mm plus 0.5mm minus 0.5mm}}
\newcommand{\spaceBelowSmallTable}[0]{\vspace{-1.5mm plus 0.5mm minus 0.5mm}}
\newcommand{\spaceBelowLargeTable}[0]{\vspace{-1mm plus 0.5mm minus 0.5mm}}
\newcommand{\spaceAboveSubfigureCaption}[0]{\vspace{-0.6\baselineskip}}
\newcommand{\spaceabovefigurecaption}[0]{\vspace{-1.5mm plus 0.5mm minus 0.5mm}}
\newcommand{\spaceBelowSmallFigure}[0]{\vspace{-3.5mm plus 0.5mm minus 0.5mm}}
\newcommand{\spaceBelowLargeFigure}[0]{\vspace{-2.5mm plus 0.5mm minus 0.5mm}}
\newcommand{\spaceAroundProof}[0]{\vspace{-1mm plus 0.5mm minus 0.5mm}}

\usepackage{xcolor}

\newcommand{\header}[1]{\textbf{#1.}}
\newcommand{\myTag}[1]{\underline{\smash{#1}}}
\newcommand{\myQuad}[0]{\hskip0.5mm}

\theoremstyle{plain}
\newtheorem{definition}{Definition}
\newtheorem{lemma}{Lemma}

\newtheorem{factor}{Distinguishing Factor}
\newtheorem{painPoint}{Pain Point}

\theoremstyle{definition}
\newtheorem{observation}{Observation}

\newcommand{\cat}[0]{\mathbin\Vert}
\newcommand{\concat}{%
  \mathop{ 
    \mathchoice{\doconcat\Large}
               {\Vert\large}
               {\Vert\normalsize}
               {\Vert\small}
    }\displaylimits 
}
\newcommand{\doconcat}[1]{%
  \vcenter{#1\kern.2ex\hbox{$\Vert$}\kern.2ex}}

\newcommand{\hide}[1]{}

\newcommand{\gold}[1]{\cellcolor{green}{#1}}
\newcommand{\silver}[1]{\cellcolor{green!40}{#1}}
\newcommand{\bronze}[1]{\cellcolor{green!20}{#1}}
\newcommand{\lose}[1]{\cellcolor{red!15}{#1}}
\newcommand{\oom}[1]{\cellcolor{red!40}{#1}}

\newcommand{\goldtwo}[1]{\cellcolor{green!50}{\underline{#1}}}
\newcommand{\silvertwo}[1]{\cellcolor{green!20}{#1}}

\newcommand{\codeurl}{\url{https://github.com/mengchillee/SlimG}\xspace}


%% file: math_commands.tex
\usepackage{amsmath,amsfonts,bm}

\def\eqref#1{equation~\ref{#1}}
\def\Eqref#1{Equation~\ref{#1}}

\def\1{\bm{1}}

\DeclareMathAlphabet{\mathsfit}{\encodingdefault}{\sfdefault}{m}{sl}
\SetMathAlphabet{\mathsfit}{bold}{\encodingdefault}{\sfdefault}{bx}{n}

\DeclareMathOperator*{\argmax}{arg\,max}

%% file: 000abstract.tex
How can we solve semi-supervised node classification in various graphs possibly with noisy features and structures?
Graph neural networks (GNNs) have succeeded in many graph mining tasks, but their generalizability to various graph scenarios is limited due to the difficulty of training, hyperparameter tuning, and the selection of a model itself.
Einstein said that we should ``make everything as simple as possible, but not simpler.''
We rephrase it into the \principle principle: a carefully-designed simple model can surpass sophisticated ones in real-world graphs.
Based on the principle, we propose \method for semi-supervised node classification, which exhibits four desirable properties:
It is (a) {\em accurate}, winning or tying on 10 out of 13 real-world datasets;
(b) {\em robust}, being the only one that handles all scenarios of graph data (homophily, heterophily, random structure, noisy features, etc.);
(c) {\em fast and scalable}, showing up to \textbf{18$\times$} faster training in million-scale graphs;
and (d) {\em interpretable}, thanks to the linearity and sparsity.
We explain the success of \method through a systematic study of the designs of existing GNNs, sanity checks, and comprehensive ablation studies.

%% file: 010intro.tex
\section{Introduction}
\label{sec:intro}

How can we solve semi-supervised node classification in various types of graphs possibly with noisy features and structures?
Graph neural networks (GNNs) \citep{Kipf17GCN, Hamilton17SAGE, Gilmer17quantum, Velickovic18GAT} have succeeded in various graph mining tasks such as node classification, clustering, or link prediction.
However, due to the difficulty of training, hyperparameter tuning, and even the selection of a model itself, many GNNs fail to show their best performance when applied to a large testbed that contains real-world graphs with various characteristics.
Especially when a graph contains noisy observations in its features and/or its graphical structure, which is common in real-world data, existing models easily overfit their parameters to such noises.

\begin{figure}
    \centering
    \vspace{2mm}
    \includegraphics[width=0.47\textwidth]{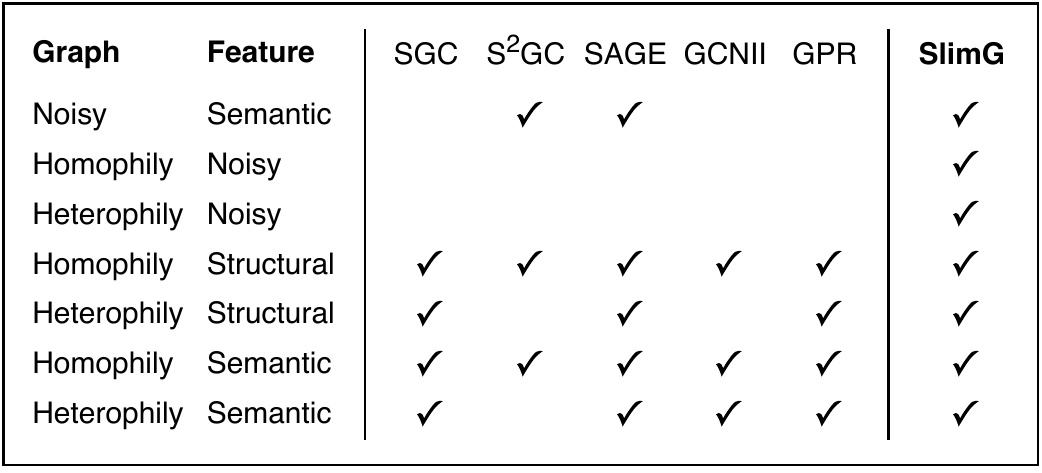}
    \spaceabovefigurecaption
    \caption{
        \myTag{\method wins on all sanity checks.}
        Each row is a specific scenario of graph data that we propose for comprehensive evaluation in Section \ref{sec:unit-tests}.
        The table is generated from the actual accuracy in Table~\ref{table:sanity-tests}: $\checkmark$ means the accuracy $\geq 80\%$.
        }
    \label{fig:overview-2}
\spaceBelowSmallFigure
\end{figure}

In response to the question, we propose \method, our novel classifier model on graphs based on the \principleLong principle: a simple carefully-designed model can be more accurate than complex ones thanks to better generalizability, robustness, and easier training.
The four design decisions of \method (D1-4 in Sec. \ref{sec:proposed}) are carefully made to follow this principle by observing and addressing 
the pain points of existing GNNs; we generate and combine various types of graph-based features (D1), design structure-only features (D2), remove redundancy in feature transformation (D3), and make the propagator function contain no hyperparameters (D4).

\begin{figure*}
    \centering
    \includegraphics[height=1.73in]{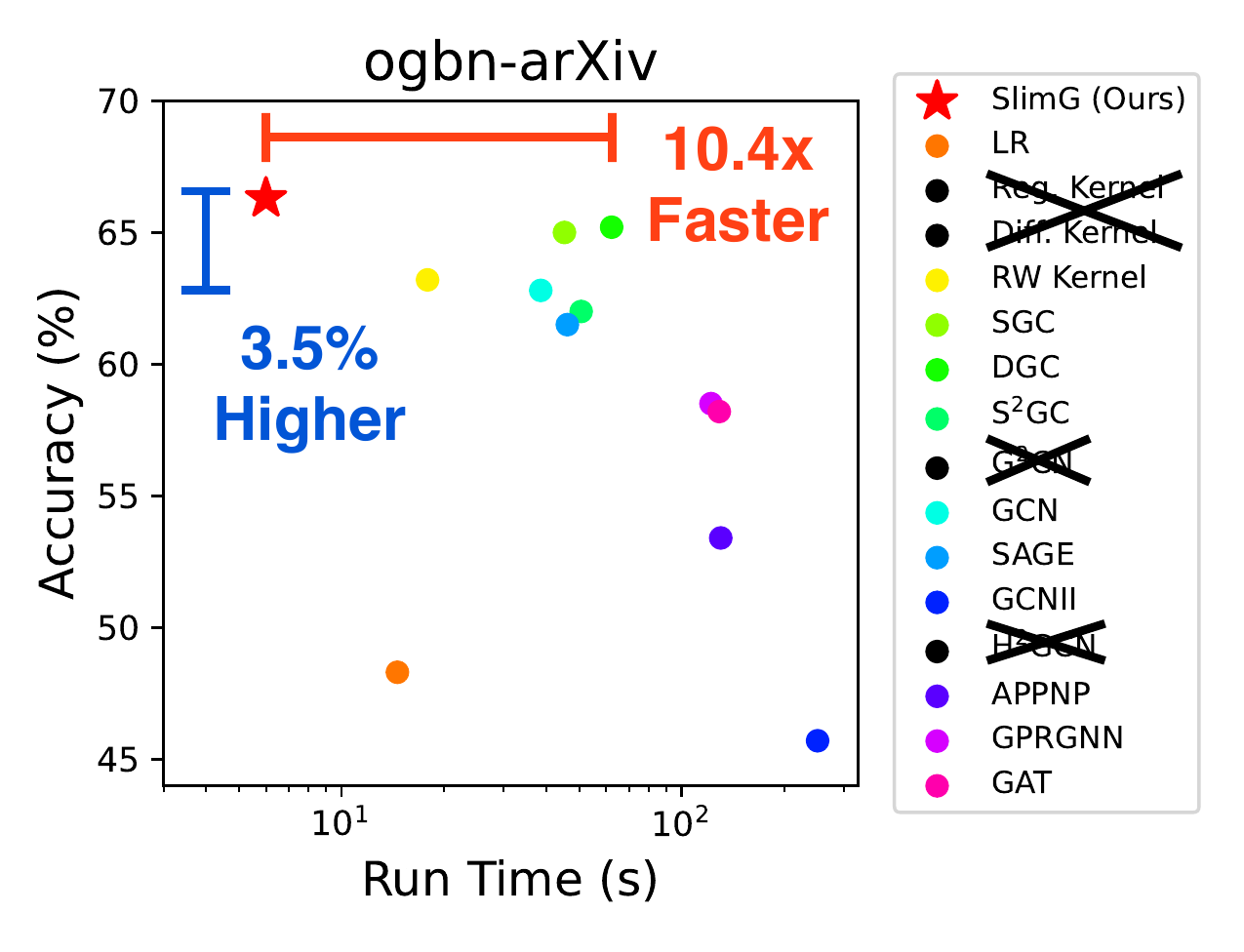}
    \hspace{-3mm}
    \includegraphics[height=1.73in]{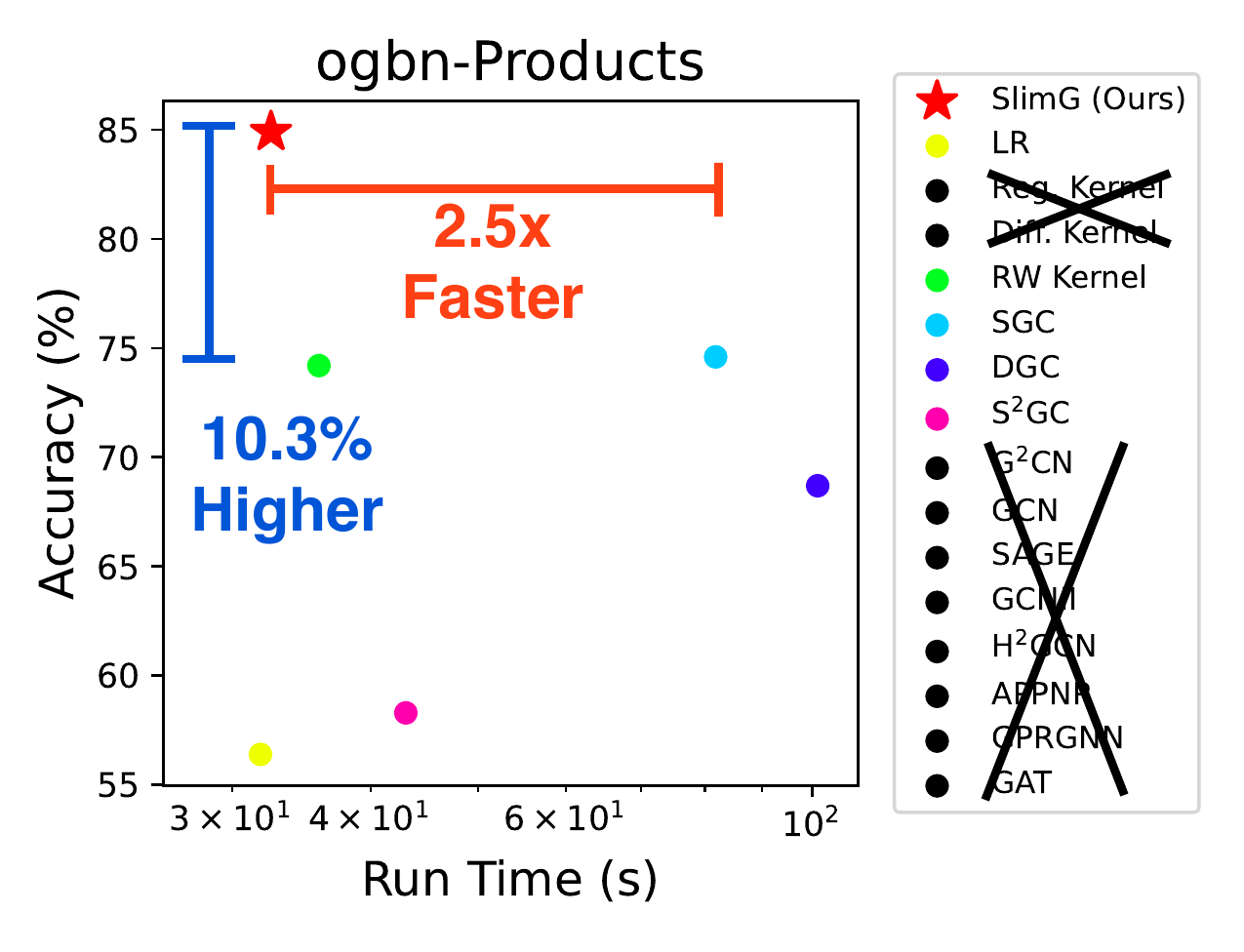}
    \hspace{-3mm}
    \includegraphics[height=1.73in]{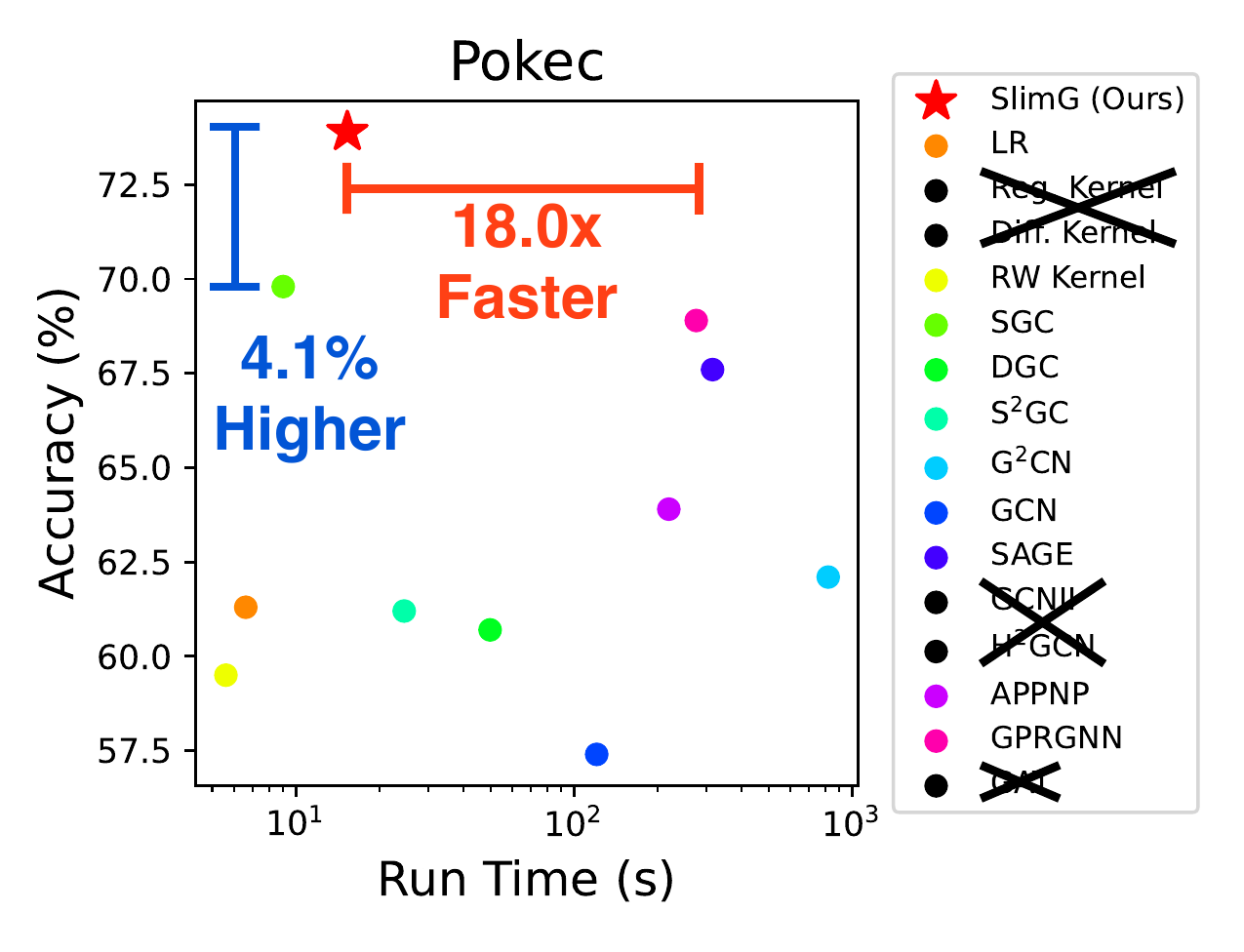}
    \spaceabovefigurecaption
    \vspace{-2mm}
    \caption{
        \myTag{\method wins both on accuracy 
            and training time} on (left) ogbn-arXiv, (middle) ogbn-Products, and (right) Pokec, which are large real-world graphs (1.2M, 61.9M, and 30.6M edges, resp.).
        Several baselines run out of memory (crossed out).
    }
    \label{fig:overview-0}
\spaceBelowLargeFigure
\end{figure*}

The resulting model, \method, is our main contribution (C1) which exhibits the following desirable properties:

\begin{compactitem}
    \item \mycontribution{C1.1 - Accurate}
        on both real-world and synthetic datasets, almost always winning or tying in the first place (see Figure~\ref{fig:overview-0}, Table~\ref{table:sanity-tests}, and Table \ref{table:expr}).
    \item \mycontribution{C1.2 - Robust},
        being able to handle numerous real settings such as homophily, heterophily, no network effects, useless features (see Figure~\ref{fig:overview-2} and Table~\ref{table:sanity-tests}).
    \item \mycontribution{C1.3 - Fast and scalable},
        using carefully chosen features, it takes only $32$ seconds on million-scale real-world graphs (ogbn-Products) on a stock server (see Figure~\ref{fig:overview-0}).
    \item \mycontribution{C1.4 - Interpretable},
        learning the largest weights on informative features, ignoring noisy ones, based on the linear decision function (see Figure~\ref{fig:interpret}).
\end{compactitem}

Not only we propose a carefully designed, effective method (in Sec. \ref{sec:proposed}), but we also explain the reasons for its success.
This is thanks to our three additional contributions (C2-4):
\begin{compactitem}
    \item \mycontribution{C2 - Explanation (Sec. \ref{sec:linearization}):}
        We propose \framework, a framework for the systematic linearization of a GNN.
        As shown in Table~\ref{table:lin-gnn}, \framework highlights the similarities, differences, and weaknesses of successful GNNs.
    \item \mycontribution{C3 - Sanity checks (Sec.~\ref{sec:unit-tests}):}
        We propose seven possible scenarios of graphs (homophily, heterophily, no network effects, etc.), which reveal the strong and weak points of each GNN; see Figure~\ref{fig:overview-2} with more details in Table~\ref{table:sanity-tests}.
    \item \mycontribution{C4 - Experiments (Sec.~\ref{sec:exp}):}
        We conduct extensive experiments to better understand the success of \method even with its simplicity.
    Our results in Tables~\ref{table:ablation-components} to \ref{table:ablrf} show that \method effectively selects the most informative component in each dataset, fully exploiting its robustness and generality.
\end{compactitem}

\begin{figure}
    \centering
    \includegraphics[width=0.45\textwidth]{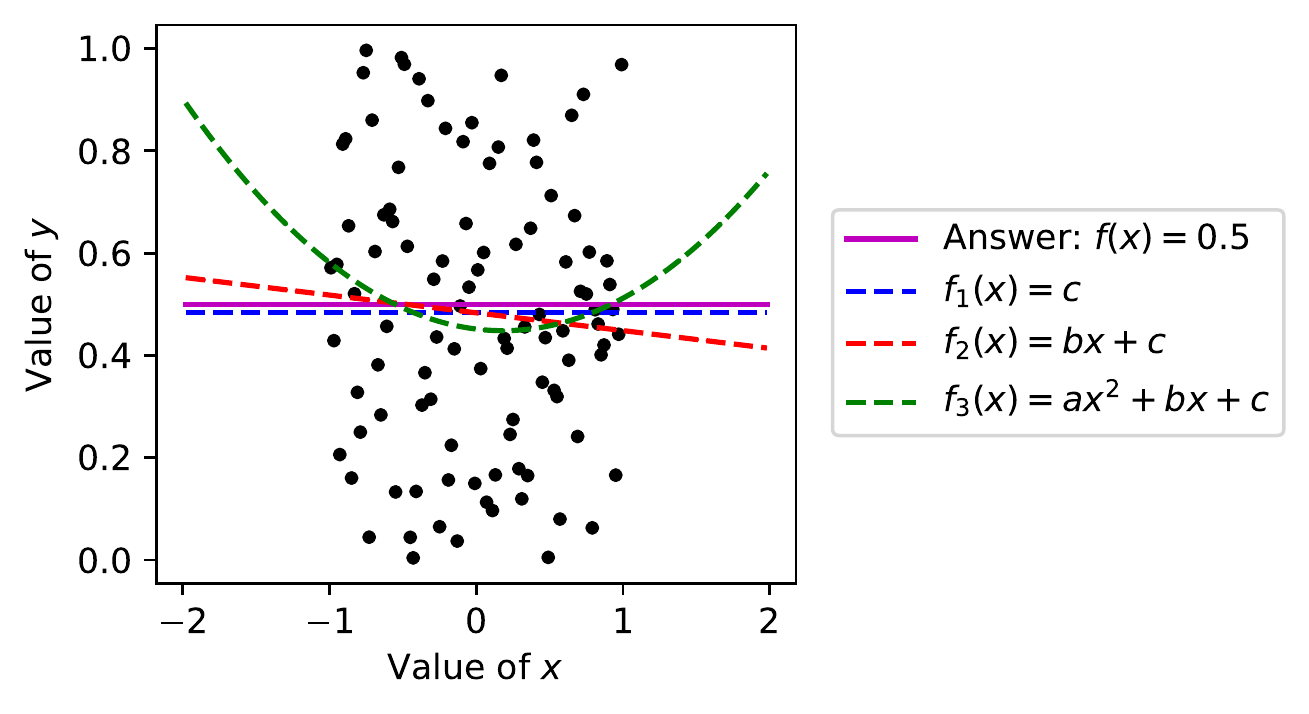}
    \spaceabovefigurecaption
    \caption{
        \myTag{Why `less is more':}
        The simple model $f_0(x)=c$ (in blue) matches the reality (in solid purple), while richer models with more polynomial powers end up capturing noise in the given data: the tiny downward trend by $f_1(x)$ (in red) and the spurious curvature by $f_2(x)$ (in green).
    }
    \label{fig:example}
\spaceBelowSmallFigure
\end{figure}

\header{Less is more}
Our justification for the counter-intuitive success
of simplicity is illustrated in Figure~\ref{fig:example}: A set of points are uniformly distributed in $x \in (-1, 1)$ and $y \in (0, 1)$, and the fitting polynomials $f_i(x)$ with degree $i=0$, 1 and 2 are given.
Notice that the simplest model $f_0$ (blue line) matches the true generator $f(x) = 0.5$.
Richer models use the 1st and the 2nd degree powers 
({\em many cooks spoil the broth})
and end up modeling tiny artifacts, like the small downward slope
of $f_1$ (red line), and the  curvature
of $f_2$ (green line).
This `many cooks' issue is more subtle and counter-intuitive than overfitting, as $f_2$ and $f_3$ have only 2 to 3 unknown parameters to fit to the hundred data points: even a \emph{small} statistical can fail if it is not matched with the underlying data-generating mechanism.

\header{Reproducibility}
Our code, along with our datasets for \emph{sanity checks}, is available at \codeurl.

%% file: 020related.tex
\section{Background and Related Works}
\label{sec:related-works}

\subsection{Background}

We define semi-supervised node classification as follows:
\begin{compactitem}
    \item \textbf{Given} An undirected graph $G = (\mathbf{A}, \mathbf{X})$, where $\mathbf{A} \in \mathbb{R}^{n \times n}$ is an adjacency matrix, $\mathbf{X} \in \mathbb{R}^{n \times d}$ is a node feature matrix, $n$ is the number of nodes, and $d$ is the number of features.
    \item \textbf{Given} The labels $\mathbf{y} \in \{1, \cdots, c\}^m$ of $m$ nodes in $G$, where $m \ll n$, and $c$ is the number of classes.
    \item \textbf{Predict} The unknown classes of $n - m$ test nodes.
\end{compactitem}

We use the following symbols to represent adjacency matrices with various normalizations and/or self-loops.
\smash{$\tilde{\mathbf{A}} = \mathbf{A} + \mathbf{I}$} is the adjacency matrix with self-loops.
\smash{$\tilde{\mathbf{D}} = \mathrm{diag}(\tilde{\mathbf{A}} \mathbf{1}_{n \times 1})$} is the diagonal degree matrix of \smash{$\tilde{\mathbf{A}}$}, where $\mathbf{1}_{n \times 1}$ is the matrix of size $n \times 1$ filled with ones.
\smash{$\AsymK{} = \smash{\tilde{\mathbf{D}}^{-1/2}} \tilde{\mathbf{A}} \tilde{\mathbf{D}}^{-1/2}$} is the symmetrically normalized \smash{$\tilde{\mathbf{A}}$}.
Similarly, \smash{$\mathbf{A}_\mathrm{sym} = \mathbf{D}^{-1/2} \mathbf{A} \mathbf{D}^{-1/2}$} is also the symmetrically normalized $\mathbf{A}$ but without self-loops.
We also use a different type of normalization \smash{$\mathbf{A}_\mathrm{row} = \mathbf{D}^{-1} \mathbf{A}$} (and accordingly \smash{$\tilde{\mathbf{A}}_\mathrm{row}$}), which we call row normalization, based on the position of the matrix $\mathbf{D}$.

As a background, we define logistic regression (LR) as a function to find the weight matrix $\mathbf{W}$ that best maps given features to labels with a linear function.

\begin{definition}[LR]
    Given a feature $\mathbf{X} \in \mathbb{R}^{n \times d}$ and a label $\mathbf{y} \in \mathbb{R}^m$, where $m$ is the number of observations such that $m \leq n$, let $\mathbf{Y} \in \mathbb{R}^{m \times c}$ be the one-hot representation of $\mathbf{y}$, and $y_{ij}$ be the $(i, j)$-th element in $\mathbf{Y}$.
    Then, logistic regression (LR) is a function that finds an optimal weight matrix $\mathbf{W} \in \mathbb{R}^{d \times c}$ from $\mathbf{X}$ and $\mathbf{y}$ as follows:
    \begin{equation}
        \mathrm{LR}(\mathbf{X}, \mathbf{y}) = \argmax_\mathbf{W} \sum_{i=1}^m \sum_{j=1}^c y_{ij} \log \mathrm{softmax}_j(\mathbf{W}^\top \mathbf{x}_i),
    \end{equation}
    where $\mathrm{softmax}_j(\cdot)$ represents selecting the $j$-th element of the result of the softmax function.
    We omit the bias term for brevity.
  \label{def:logistic-regression}
\end{definition}

\subsection{Related Works}

We introduce related works categorized into graph neural networks (GNN), linear GNNs, and graph kernel methods.

\header{Graph neural networks}
There exist many recent GNN variants; recent surveys \citep{Zhou20survey, Wu21survey} group them into spectral models \citep{Defferrard16ChebNet, Kipf17GCN}, sampling-based models \citep{Hamilton17SAGE, Ying18web, Zhu20H2GCN}, attention-based models \citep{Velickovic18GAT, Kim21SuperGAT, Brody22GATv2}, and deep models with residual connections \citep{Li19DeepGCN, Chen20GCNII}.
Decoupled models \citep{Klicpera19APPNP, Klicpera19GDC, Chien21GPRGNN} separate the two major functionalities of GNNs: the node-wise feature transformation and the propagation.
GNNs are often fused with graphical inference \citep{Yoo19BPN, Huang21CnS} to further improve the predicted results.
These GNNs have shown great performance in many graph mining tasks, but suffer from limited robustness when applied to graphs with various characteristics possibly having noisy observations, especially in semi-supervised learning.

\header{Linear graph neural networks}
\citet{Wu19SGC} proposed SGC by removing the nonlinear activation functions of GCN \citep{Kipf17GCN}, reducing the propagator function to a simple matrix multiplication.
\citet{Wang21DGC} and \citet{Zhu21S2GC} improved SGC by manually adjusting the strength of self-loops with hyperparameters, increasing the number of propagation steps.
\citet{Li22G2CN} proposed G$^2$CN, which improves the accuracy of DGC \citep{Wang21DGC} on heterophily graphs by combining multiple propagation settings (i.e. bandwidths).
The main limitation of these models is the high complexity of propagator functions with many hyperparameters, which impairs both the robustness and interpretability of decisions even with linearity.

\header{Graph kernel methods}
Traditional works on graph kernel methods \citep{Smola03Kernels, Ioannidis17Kernels} are closely related to linear GNNs, which can be understood as applying a linear graph kernel to transform the raw features.
A notable limitation of such kernel methods is that they are not capable of addressing various scenarios of real-world graphs, such as heterophily graphs, as their motivation is to aggregate all information in the local neighborhood of each node, rather than ignoring noisy and useless ones.
We implement three popular kernel methods as additional baselines and show that our \method outperforms them in both synthetic and real graphs.

%% file: 030linear.tex
\section{Proposed Framework: \framework}
\label{sec:linearization}

Why do GNNs work well when they do? In what cases will a GNN fail?
We answer these questions with \framework, our proposed framework for revealing the essence of each GNN.
The idea is to derive the essential \emph{feature propagator} function on which each variant is based, ignoring nonlinearity, so that all models are comparable on the same ground.
The observations from \framework motivate us to propose our method \method, which we describe in Section \ref{sec:proposed}.

\begin{definition}[Linearization]
	Given a graph $G = (\mathbf{A}, \mathbf{X})$, let $f(\cdot; \theta)$ be a node classifier function to predict the labels of all nodes in $G$ as $\hat{\mathbf{y}} = f(\mathbf{A}, \mathbf{X}; \theta)$, where $\theta$ is the set of parameters.
	Then, $f$ is \emph{linearized} if $\theta = \{ \mathbf{W} \}$ and the optimal weight matrix \smash{$\mathbf{W}^* \in \mathbb{R}^{h \times c}$} is given as
	\begin{equation}
		\mathbf{W}^* = \mathrm{LR}(\mathcal{P}(\mathbf{A}, \mathbf{X}), \mathbf{y}),
	\label{eq:linear-gnn}
	\end{equation}
	where $\mathcal{P}$ is a feature propagator function that is linear with $\mathbf{X}$ and contains no learnable parameters, and $\mathcal{P}(\mathbf{A}, \mathbf{X}) \in \mathbb{R}^{n \times h}$.
	We ignore the bias term without loss of generality.
\label{def:linear-gnn}
\end{definition}

\begin{definition}[\framework]
    Given a GNN $f$, \framework is to represent $f$ as a linearized GNN by replacing all (nonlinear) activation functions in $f$ with the identity function and deriving a variant $f'$ that is at least as expressive as $f$ but contains no parameters in $\mathcal{P}$.
\label{def:linearization}
\end{definition}

\framework represents the characteristic of a GNN as a linear feature propagator function $\mathcal{P}$, which transforms raw features $\mathbf{X}$ by utilizing the graph structure $\mathbf{A}$.
Lemma \ref{lemma:linear-models} shows that \framework generalizes existing linear GNNs.
Logistic regression is also represented by \framework with the identity propagator $\mathcal{P}(\mathbf{A}, \mathbf{X}) = \mathbf{X}$.

\begin{lemma}
    \framework includes existing linear graph neural networks as its special cases: SGC, DGC, S$^2$GC, and G$^2$CN.
    \label{lemma:linear-models}
\end{lemma}

\begin{proof}
\spaceAroundProof
The proof is given in Appendix \ref{appendix:proof-linear-models}.
\spaceAroundProof
\end{proof}


\begin{table}
\centering
\caption{
    \myTag{\framework encompasses popular GNNs.}
    The * and ** superscripts mark fully and partially linearized models, respectively. 
    We derive Pain Points (Sec. \ref{ssec:pain-points}) and Distinguishing Factors (Sec. \ref{ssec:linear-observations}) of the variants through \framework.
}
\spaceBelowTableCaption
\resizebox{0.47\textwidth}{!}{
\begin{tabular}{l|l|r}
    \toprule
    Model & Type & Propagator function $\mathcal{P}(\mathbf{A}, \mathbf{X})$ \\
    \midrule
    LR & Linear &
        $\mathbf{X}$ \\
    \midrule
    SGC & Linear &
        $\AsymK{K} \myQuad \mathbf{X}$ \\
    DGC & Linear &
        $[(1 - T/K) \mathbf{I} + (T/K) \AsymK{}]^K \myQuad \mathbf{X}$ \\
    S$^2$GC & Linear &
        $\smash{\sum_{k=1}^K} (\alpha \mathbf{I} + (1 - \alpha) \AsymK{k}) \myQuad \mathbf{X}$ \\
    G$^2$CN & Linear &
        $\concat_{i=1}^N [\mathbf{I} - (T_i/K) ((b_i - 1) \mathbf{I} + \mathbf{A}_\mathrm{sym})^2]^K \myQuad \mathbf{X}$ \\
    \midrule
    PPNP* & Decoupled &
        $(\mathbf{I} - (1 - \alpha) \AsymK{})^{-1} \myQuad \mathbf{X}$ \\
    APPNP* & Decoupled &
        $[ \sum_{k=0}^{K-1} \alpha (1 - \alpha)^k \AsymK{k} + (1 - \alpha)^K \AsymK{K} ] \myQuad \mathbf{X}$ \\
    GDC* & Decoupled &
        $\mathbf{S} = \mathrm{sparse}_\epsilon(\sum_{k=0}^\infty (1 - \alpha)^k \tilde{\mathbf{A}}_\mathrm{sym}^k)$ for $\tilde{\mathbf{S}}_\mathrm{sym} \myQuad \mathbf{X}$ \\
    GPR-GNN* & Decoupled &
        $\concat_{k=0}^K \AsymK{k} \myQuad \mathbf{X}$ \\
    \midrule
    ChebNet* & Coupled &
        $\concat_{k=0}^{K-1} \mathbf{A}_\mathrm{sym}^k \myQuad \mathbf{X}$ \\
    GCN* & Coupled &
        $\AsymK{K} \myQuad \mathbf{X}$ \\
    SAGE* & Coupled & 
        $\concat_{k=0}^K \mathbf{A}_\mathrm{row}^k \myQuad \mathbf{X}$ \\
    GCNII* & Coupled & 
        $\concat_{k=0}^{K-2} \AsymK{k} \mathbf{X} \cat ((1 - \alpha) \AsymK{K} + \alpha \AsymK{K-1}) \myQuad \mathbf{X}$ \\
    H$_2$GCN* & Coupled & 
        $\concat_{k=0}^{2K} \mathbf{A}_\mathrm{sym}^k \mathbf{X}$ \\
    \midrule
    GAT** & Attention &
        $\prod_{k=1}^K [ \mathrm{diag}(\mathbf{X} \mathbf{w}_{k,1}) \tilde{\mathbf{A}} + \tilde{\mathbf{A}} \mathrm{diag}(\mathbf{X} \mathbf{w}_{k,2}) ] \myQuad \mathbf{X}$ \\
    DA-GNN** & Attention &
        $\sum_{k=0}^K \mathrm{diag}(\AsymK{k} \mathbf{X} \mathbf{w}) \AsymK{k} \myQuad \mathbf{X}$ \\
    \bottomrule
\end{tabular}
}
\label{table:lin-gnn}
\spaceBelowSmallTable
\end{table}

In Table \ref{table:lin-gnn}, we conduct a comprehensive linearization of existing GNNs using \framework to understand the fundamental similarities and differences among GNN variants.
The models are categorized into linear, decoupled, coupled, and attention models.
We ignore bias terms for simplicity, without loss of generality.
Refer to Appendix \ref{appendix:linearization} for details of the linearization process for each model.


\subsection{Pain Points of Existing GNNs}
\label{ssec:pain-points}

Based on the comprehensive linearization in Table \ref{table:lin-gnn}, we derive four pain points of existing GNNs which we address in Section \ref{sec:proposed}.

\begin{painPoint}[Lack of robustness]
    All models in Table~\ref{table:lin-gnn} fail to handle multiple graph scenarios at the same time, i.e., graphs with homophily, heterophily, no network effects, or useless features.
\label{pp:generality}
\end{painPoint}

Most models in Table~\ref{table:lin-gnn} make an implicit assumption on given graphs, such as homophily or heterophily, rather than being able to perform well in multiple scenarios at the same time.
For example, all models except ChebNet, SAGE, and H$_2$GCN have self-loops in the new adjacency matrix, emphasizing the local neighborhood of each node even in graphs with heterophily or no network effects.
This is the pain point that we also observe empirically from the sanity checks (in Table~\ref{table:sanity-tests}), where none of the existing models succeeds in making reasonable accuracy in all cases of synthetic graphs.

\begin{painPoint}[Vulnerability to noisy features]
    All models in Table~\ref{table:lin-gnn} cannot fully exploit the graph structure if the features are noisy, since they depend on the node feature matrix $\mathbf{X}$.
\label{pp:noisyFeatures}
\end{painPoint}

If a graph does not have a feature matrix $\mathbf{X}$, a common solution is to introduce one-hot features \cite{Kipf17GCN}, i.e., $\mathbf{X} = \mathbf{I}$, although it increases the running time of the model a lot.
On the other hand, if $\mathbf{X}$ exists but some of its elements are meaningless with respect to the target classes, models whose propagator functions rely on $\mathbf{X}$ suffer from the noisy elements.
In such cases, a desirable property for a model is to adaptively emphasize important features or disregard noisy ones to maximize its generalization performance, which is not satisfied by any of the existing models in Table~\ref{table:lin-gnn}.

\begin{painPoint}[Efficiency and effectiveness]
    Concatenation-based models in Table \ref{table:lin-gnn} create spurious correlations between feature elements, requiring more parameters than in other models.
\label{pp:efficiency}
\end{painPoint}

Existing models such as GPR-GNN, SAGE, and GCNII in Table~\ref{table:lin-gnn} perform the concatenation of multiple feature matrices transformed in different ways.
For example, GPR-GNN concatenates \smash{$\AsymK{k} \mathbf{X}$} for different values of $k$ from 0 to $K$, where $K$ is a hyperparameter.
Such a concatenation-based propagation limits the efficiency of a model in two ways.
First, this increases the number of parameters $K$ times, since the model needs to learn a separate weight matrix for each given feature matrix.
Second, this creates spurious correlations in the resulting features, since the feature matrices like \smash{$\tilde{\mathbf{A}}_\mathrm{sym} \mathbf{X}$} and \smash{$\AsymK{2} \mathbf{X}$} have high correlations with each other.

\begin{painPoint}[Many hyperparameters]
    Hyperparameters in $\mathcal{P}$ impair its interpretability and require extensive tuning.
\label{pp:hyperparameters}
\end{painPoint}

Most models in Table \ref{table:lin-gnn}, even the linear models such as DGC and G$^2$CN, contain many hyperparameters in the propagator function $\mathcal{P}$.
Such hyperparameters lead to two limitations.
First, the interpretability of the weight matrix $\mathbf{W}$ is impaired, since it is learned on top of the transformed feature $\mathcal{P}(\mathbf{A}, \mathbf{X})$ whose meaning changes arbitrarily by the choice of its hyperparameters.
For example, DGC changes the number $K$ of propagation steps between 250 and 900 in real-world datasets, making it hard to have consistent observations from the generated features.
Second, $\mathcal{P}(\mathbf{A}, \mathbf{X})$ should be computed for every new choice of hyperparameters, while it can be cached and reused for searching hyperparameters outside $\mathcal{P}$.

\subsection{Distinguishing Factors}
\label{ssec:linear-observations}

What potential choices do we have in designing a general approach that addresses the pain points?
We analyze the fundamental similarities and differences among the GNN variants in Table \ref{table:lin-gnn}.

\begin{factor}[Combination of features]
    How should we combine node features, the immediate neighbors' features, and the $K$-step-away neighbors' features?
\end{factor}

GNNs propagate information by multiplying the feature $\mathbf{X}$ with (a variant of) the adjacency matrix $\mathbf{A}$ multiple times.
There are two main choices in Table \ref{table:lin-gnn}: \textbf{(1)} the summation of transformed features (most models), and \textbf{(2)} the concatenation of features (GPR-GNN, GraphSAGE, GCNII, and H$_2$GCN).
Simple approaches like SGC are categorized as the summation due to the self-loops in \smash{$\tilde{\mathbf{A}}_\mathrm{sym}$}.

\begin{factor}[Modification of $\mathbf{A}$]
    How should we normalize or modify the adjacency matrix $\mathbf{A}$?
\end{factor}

The three prevailing choices are given as follows: \textbf{(1)} symmetric vs. row normalization, \textbf{(2)} the strength of self-loops, including making zero self-loops, and \textbf{(3)} static vs. dynamic adjustment based on the given features.
Most models use the symmetric normalization $\smash{\AsymK{}}$ with self-loops, but some variants avoid self-loops and use either row normalization $\mathbf{A}_\mathrm{row}$ or symmetric one $\smash{\mathbf{A}_\mathrm{sym}}$.
Recent models such as DGC, G$^2$CN, and GCNII determine the weight of self-loops with hyperparameters, since strong self-loops allow distant propagation with a large value of $K$.
Finally, attention-based models learn the elements in $\mathbf{A}$ dynamically based on node features, making propagator functions quadratic with $\mathbf{X}$, not linear.

\begin{factor}[Heterophily]
    What to do if the direct neighbors differ in their features or labels?
\end{factor}

In such cases, the simple aggregation of the features of immediate neighbors may hurt performance, and therefore, several GNNs do suffer under heterophily as shown in Table~\ref{table:sanity-tests}.
GNNs that can handle heterophily adopt one or more of these ideas:
\textbf{(1)} using the square of $\mathbf{A}$ as the base structure (G$^2$CN);
\textbf{(2)} learning different weights for different steps (GPR-GNN, ChebNet, SAGE, and GCNII), and
\textbf{(3)} making small or no self-loops in the modification of $\mathbf{A}$ (DGC, S$^2$GC, G$^2$CN, and H$^2$GCN).
The idea is to avoid or downplay the effect of immediate (and odd-step-away) neighbors.
Self-loops hurt under heterophily, as they force to have information of all intermediate neighbors by acting as the implicit summation of transformed features.

%% file: 040method.tex
\section{Proposed Method: \method}
\label{sec:proposed}

We propose \method, a novel method that addresses the limitations of existing models with strict adherence to the \principleLong principle.
We first derive four \emph{design decisions} (D1-D4) that directly address the pain points (PPs) of existing GNN models and propose the following propagator function of \method:
\begin{equation}
\boxed{
    \mathcal{P}(\mathbf{A}, \mathbf{X}) = 
        \underbrace{\mathbf{U}}_{\text{Structure}} \cat
        \underbrace{g(\mathbf{X})}_{\text{Features}} \cat
        \underbrace{g(\mathbf{A}_\mathrm{row}^2 \mathbf{X})}_{\text{2-step neighbors}} \cat
        \underbrace{g(\tilde{\mathbf{A}}_\mathrm{sym}^2 \mathbf{X})}_{\text{Neighbors}}
}
\label{eq:proposed}
\end{equation}
where $g(\cdot)$ is the principal component analysis (PCA) for the orthogonalization of each component, followed by an L2 normalization, and $\mathbf{U} \in \mathbb{R}^{n \times r}$ contains $r$-dimensional structural features independent of node features $\mathbf{X}$, derived by running the low-rank singular value decomposition (SVD) on the adjacency matrix $\mathbf{A}$.

\paragraph{\underline{\smash{\textbf{D1}: Concatenating winning normalizations}} (for PP \ref{pp:generality} - robustness)}
The main principle of \method to acquire robustness and generalizability, in response to Pain Point \ref{pp:generality}, is to transform the raw features into various forms and then combine them through concatenation.
In this way, \method is able to emphasize essential features or ignore useless ones by learning separate weights for different components.
The four components of \method in \Eqref{eq:proposed} show their strength in different cases: structural features $\mathbf{U}$ for graphs with noisy features, self-features $\mathbf{X}$ for a noisy structure, two-step aggregation $\mathbf{A}^2_\mathrm{row}$ for heterophily graphs, and smoothed two-hop aggregation \smash{$\tilde{\mathbf{A}}^2_\mathrm{sym}$} of the local neighborhood for homophily.

Specifically, we use the row-normalized matrix $\mathbf{A}_\mathrm{row}$ with no self-loops due to the limitations of the symmetric normalization \smash{$\AsymK{}$}:
First, the self-loops force one to combine all intermediate neighbors of each node until the $K$-hop distance, even in heterophily graphs where the direct neighbors should be avoided.
Second, neighboring features are rescaled based on the node degrees during an aggregation, even when we want simple aggregation of $K$-hop neighbors preserving the original scale of features.
We thus use $\mathbf{A}^2_\mathrm{row}$ along with the popular transformation \smash{$\tilde{\mathbf{A}}^2_\mathrm{sym}$} in \method.

\paragraph{\underline{\smash{\textbf{D2}: Structural features}} (for PP \ref{pp:noisyFeatures} - noisy features)}

In response to Pain Point~\ref{pp:noisyFeatures}, we have to resort to the structure $\mathbf{A}$ ignoring $\mathbf{X}$ when features are missing, noisy, or useless for classification.
Thanks to our design decision (D1) for concatenating different components, we can safely add to $\mathcal{P}$ structure-based features which the model can utilize adaptively based on the amount of information $\mathbf{X}$ provides.
However, it is not effective to use raw $\mathbf{A}$, which requires the model to learn a separate weight vector for each node, severely limiting its generalizability.
We thus adopt low-rank SVD with rank $r$ to extract structural features $\mathbf{U}$.
The value of $r$ is selected to keep $90\%$ of the energy of $\mathbf{A}$, where the sum of the largest $r$ squared singular values divided by the squared Frobenius norm of $\mathbf{A}$ is smaller than $0.9$.
If the chosen value of $r$ is larger than $d$ in large graphs, we set $r$ to $d$ for the size consistency between different components.


\begin{figure}
    \centering
    \subfloat[Sanity check matrix]{
        \small
        \begin{tabular}{l|lll}
        \toprule
        \multirow{2}{*}{\textbf{Structure}} & \multicolumn{3}{c}{\textbf{Feature}} \\
        & Semantic $\mathbf{X}$ & Structural $\mathbf{X}$ & Random $\mathbf{X}$ \\
        \midrule
        Homophily $\mathbf{A}$ & Both help & Both help & $\mathbf{A}$ helps \\
        Heterophily $\mathbf{A}$ & Both help & Both help & $\mathbf{A}$ helps \\
        Uniform $\mathbf{A}$ & $\mathbf{X}$ helps & None helps & None helps \\
        \bottomrule
        \end{tabular}
        \label{fig:sanity-matrix}
    }
    
    \begin{subfigure}{0.15\textwidth}
        \includegraphics[height=1.1in]{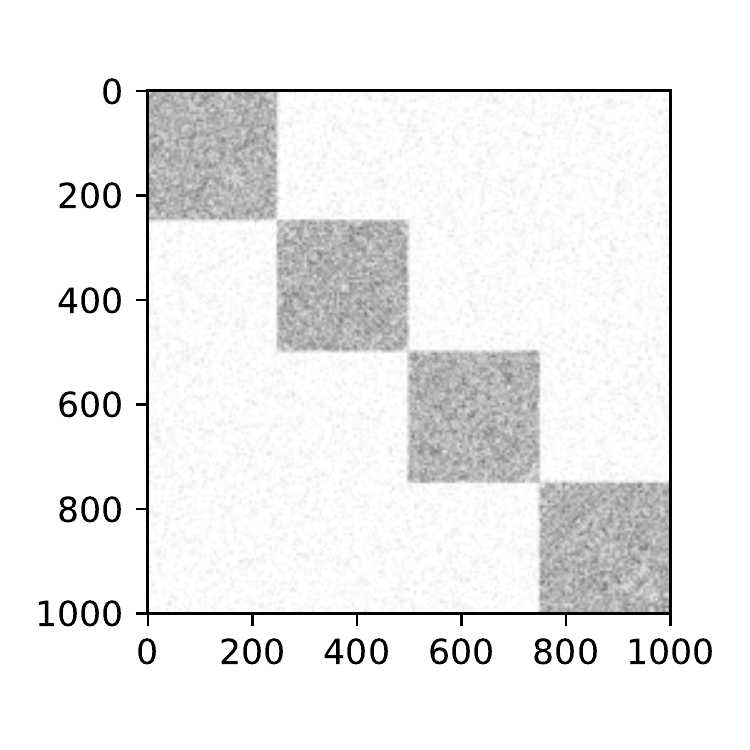}
        \vspace{-6mm}
        \caption{Homophily $\mathbf{A}$}
        \label{fig:adj-homophily}
    \end{subfigure}
    \begin{subfigure}{0.15\textwidth}
        \includegraphics[height=1.1in]{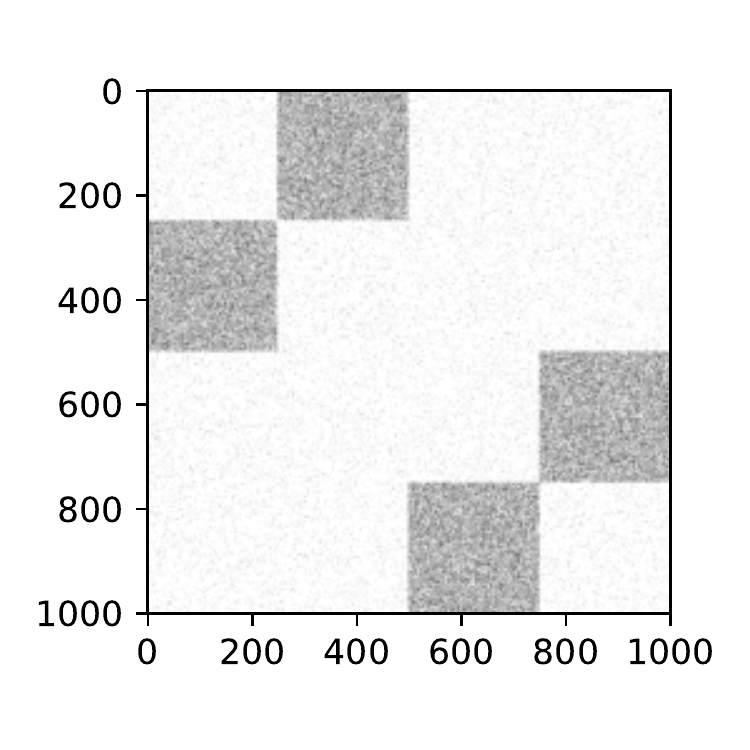}
        \vspace{-6mm}
        \caption{Heterophily $\mathbf{A}$}
        \label{fig:adj-heterophily}
    \end{subfigure}
    \begin{subfigure}{0.15\textwidth}
        \includegraphics[height=1.1in]{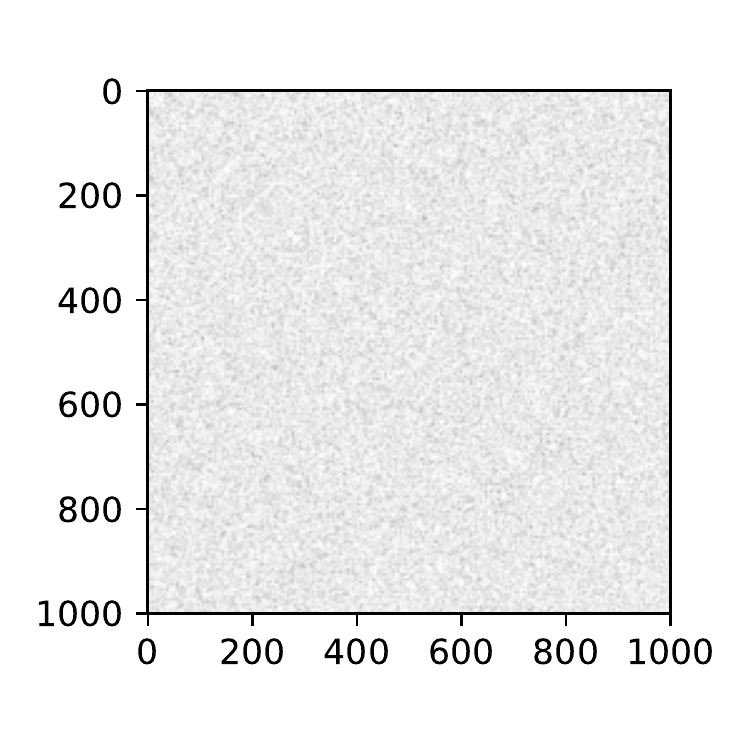}
        \vspace{-6mm}
        \caption{Uniform $\mathbf{A}$}
        \label{fig:adj-uniform}
    \end{subfigure}

    \spaceabovefigurecaption
    \caption{
        \myTag{Illustration of our sanity checks.}
        (a) We consider 3 possibilities for each of $\mathbf{A}$ and $\mathbf{X}$, creating a total of 9 cases.
        (b - d) We visualize the adjacency matrices from the three cases of $\mathbf{A}$, which exhibit different structural patterns.
    }
    \label{fig:sanity-all}
\spaceBelowSmallFigure
\end{figure}



\begin{table*}
    \renewcommand{\arraystretch}{1.1}
    \centering
    \caption{
        \myTag{\method wins on all sanity checks.}
        Each value denotes the average and the standard deviation of accuracy from five runs.
        There are three groups of scenarios: (left) only features $\mathbf{X}$ help;
        (middle) only connectivity $\mathbf{A}$ helps;
        (right) both help.
        Green (\colorbox{green}{}, \colorbox{green!40}{}, \colorbox{green!20}{}) marks the top three (higher is darker);
        red (\colorbox{red!15}{}) marks the ones that are too low (2$\sigma$ below the third place).
        \method is the only method without red cells and achieves the best average accuracy and average rank (variance in the parentheses).
    }
    \spaceBelowTableCaption
    \resizebox{0.97\textwidth}{!}{
    \begin{tabular}{l|c|cc|cccc|C{1.5cm}|r}
        \toprule
        \multirow{3}{*}{\textbf{Model}}
            & Only $\mathbf{X}$ helps
            & \multicolumn{2}{c|}{Only $\mathbf{A}$ helps}
            & \multicolumn{4}{c|}{Both $\mathbf{X}$ and $\mathbf{A}$ help} 
            & \multirow{3}{*}{\textbf{Avg. Acc}}
            & \multirow{3}{*}{\textbf{Avg. Rank}}\\
        & Semantic $\mathbf{X}$
            & Random $\mathbf{X}$
            & Random $\mathbf{X}$
            & Structural $\mathbf{X}$
            & Structural $\mathbf{X}$
            & Semantic $\mathbf{X}$
            & Semantic $\mathbf{X}$
            & & \\
        & Uniform $\mathbf{A}$
            & Homophily
            & Heterophily
            & Homophily
            & Heterophily
            & Homophily
            & Heterophily 
            & & \\
        \midrule
        LR
            & \gold{83.7$\pm$0.6}
            & \lose{24.2$\pm$0.7}
            & \lose{24.2$\pm$0.7}
            & \lose{71.4$\pm$0.9}
            & \lose{66.8$\pm$2.2}
            & \lose{83.4$\pm$0.6}
            & \lose{83.4$\pm$0.6}
            & 62.4 (26.9)
            & 10.7 (5.4) \\
        \midrule
        Reg. Kernel
            & \silver{82.7$\pm$0.5}
            & \lose{27.9$\pm$0.4}
            & \lose{24.3$\pm$1.0}
            & \lose{75.7$\pm$0.2}
            & \lose{65.3$\pm$1.6}
            & \lose{91.5$\pm$0.5}
            & \lose{79.5$\pm$0.3}
            & 63.8 (27.0)
            & 10.4 (4.3) \\
        Diff. Kernel
            & \lose{26.8$\pm$1.7}
            & \lose{38.0$\pm$8.7}
            & \lose{37.6$\pm$7.5}
            & \lose{79.5$\pm$0.3}
            & \lose{73.5$\pm$0.6}
            & \lose{70.9$\pm$23.}
            & \lose{56.1$\pm$27.}
            & 54.6 (20.7)
            & 10.6 (4.0) \\
        RW Kernel
            & \lose{72.2$\pm$0.7}
            & \lose{37.0$\pm$0.4}
            & \lose{24.5$\pm$1.3}
            & \lose{81.3$\pm$1.2}
            & \lose{51.0$\pm$1.1}
            & 94.5$\pm$0.9
            & \lose{57.8$\pm$0.7}
            & 59.8 (24.7)
            & 10.4 (3.6) \\
        \midrule
        SGC
            & \lose{44.6$\pm$9.8}
            & \lose{64.3$\pm$0.7}
            & \lose{50.2$\pm$14.}
            & 87.1$\pm$0.6
            & \lose{84.3$\pm$0.5}
            & 93.9$\pm$0.9
            & 91.5$\pm$0.5
            & \bronze{73.7 (20.4)}
            & \bronze{5.7 (3.1)} \\
        DGC
            & \lose{63.8$\pm$1.0}
            & \lose{50.5$\pm$13.}
            & \lose{26.0$\pm$0.9}
            & \bronze{88.6$\pm$1.0}
            & \lose{45.3$\pm$1.3}
            & \gold{96.2$\pm$0.4}
            & \lose{54.0$\pm$0.6}
            & 60.6 (24.6)
            & 8.3 (5.9) \\
        S$^2$GC
            & 79.9$\pm$0.6
            & \lose{38.5$\pm$12.}
            & \lose{25.4$\pm$0.9}
            & 88.4$\pm$1.0
            & \lose{67.9$\pm$1.5}
            & \silver{95.9$\pm$0.6}
            & \lose{78.0$\pm$0.5}
            & 67.7 (26.2)
            & 7.4 (3.4) \\
        G$^2$CN
            & \lose{25.2$\pm$0.3}
            & \lose{24.2$\pm$1.1}
            & \lose{25.0$\pm$0.1}
            & 88.5$\pm$1.0
            & \silver{88.6$\pm$1.2}
            & \lose{24.3$\pm$1.1}
            & \lose{50.7$\pm$31.}
            & 46.6 (30.2)
            & 11.6 (6.3) \\
        \midrule
        GCN
            & \lose{36.3$\pm$3.5}
            & \lose{46.7$\pm$8.0}
            & \lose{43.7$\pm$1.9}
            & \lose{83.3$\pm$1.3}
            & \lose{72.2$\pm$1.7}
            & \lose{91.2$\pm$1.2}
            & \lose{80.3$\pm$3.9}
            & 64.8 (22.1)
            & 8.1 (3.0) \\
        SAGE
            & 80.3$\pm$1.1
            & \lose{31.1$\pm$0.7}
            & \lose{34.6$\pm$2.1}
            & \lose{83.9$\pm$0.8}
            & \lose{81.3$\pm$0.7}
            & 94.4$\pm$0.5
            & \gold{94.4$\pm$0.9}
            & 71.4 (27.0)
            & \bronze{5.7 (2.9)} \\
        GCNII
            & \lose{73.5$\pm$1.2}
            & \lose{30.7$\pm$0.7}
            & \lose{27.1$\pm$1.3}
            & \lose{84.2$\pm$0.8}
            & \lose{69.0$\pm$1.4}
            & \lose{90.6$\pm$0.9}
            & \lose{80.4$\pm$1.2}
            & 65.1 (25.7)
            & 8.7 (1.8) \\
        H$^{2}$GCN
            & 80.2$\pm$1.5
            & \lose{27.0$\pm$1.0}
            & \lose{27.5$\pm$0.8}
            & \lose{78.0$\pm$0.9}
            & \lose{74.6$\pm$1.3}
            & 91.9$\pm$0.7
            & 92.2$\pm$0.9
            & 67.3 (28.2)
            & 8.0 (3.9) \\
        APPNP
            & \lose{66.0$\pm$2.6}
            & \lose{30.3$\pm$1.2}
            & \lose{25.2$\pm$0.7}
            & \lose{71.2$\pm$4.9}
            & \lose{43.8$\pm$2.0}
            & \lose{83.2$\pm$3.8}
            & \lose{58.7$\pm$4.5}
            & 54.1 (21.6)
            & 12.9 (2.0) \\
        GPR-GNN
            & \lose{73.4$\pm$0.4}
            & \lose{74.6$\pm$0.7}
            & \lose{65.9$\pm$2.1}
            & \gold{89.9$\pm$0.6}
            & \bronze{87.6$\pm$1.2}
            & \bronze{95.0$\pm$1.1}
            & \bronze{91.9$\pm$1.1}
            & \silver{82.6 (11.2)}
            & \silver{3.3 (2.1)} \\
        GAT
            & \lose{32.7$\pm$5.5}
            & \lose{42.6$\pm$4.8}
            & \lose{36.8$\pm$5.7}
            & \lose{64.0$\pm$5.7}
            & \lose{55.6$\pm$6.8}
            & \lose{68.5$\pm$7.1}
            & \lose{67.0$\pm$12.}
            & 52.5 (15.0)
            & 11.6 (4.1) \\
        \midrule
        \method (Ours)
            & \bronze{81.0$\pm$1.1}
            & \silver{87.1$\pm$1.4}
            & \gold{89.2$\pm$1.2}
            & 88.1$\pm$0.5
            & \gold{88.9$\pm$0.7}
            & 94.4$\pm$0.6
            & \silver{93.9$\pm$0.5}
            & \gold{88.9 (\ \ 4.5)}
            & \gold{2.6 (1.8)} \\
        \bottomrule
    \end{tabular}
    }
\label{table:sanity-tests}
\spaceBelowLargeTable
\end{table*}


\paragraph{\underline{\smash{\textbf{D3}: Orthogonalization and sparsification}} (for PP \ref{pp:efficiency} - collinearity)}

To improve the efficiency and effectiveness of \method in response to Pain Point~\ref{pp:efficiency}, we use two reliable methods to further transform the features: dimensionality reduction by PCA and regularization by group LASSO.
First, we run PCA on each of the four components to orthogonalize them and to improve the consistency of learned weights.
Second, we use group LASSO to learn sparse weights on the component level, preserving the relative magnitude of each element and suppressing noisy features.
To assure the consistency between components (especially with the structural features $\mathbf{U}$), we force all components to have the same dimensionality by selecting $r$ features from each component when adopting PCA.

\paragraph{\underline{\smash{\textbf{D4}: Multi-level neighborhood aggregation}} (for PP \ref{pp:hyperparameters} - hyperparameters)}

Our propagator function $\mathcal{P}$ considers multiple levels of neighborhoods through the concatenation of different components.
This allows us to remove all hyperparameters from $\mathcal{P}$ to tune for each dataset, in response to Pain Point~\ref{pp:hyperparameters}, gaining in both interpretability and efficiency.
Specifically, $\mathbf{X}$, \smash{$\tilde{\mathbf{A}}^2_\mathrm{sym}$}, and $\mathbf{A}^2_\mathrm{row}$ aggregate the zero-, one-, and two-hop neighborhood of each node, respectively, considering the self-loops included in \smash{$\tilde{\mathbf{A}}^2_\mathrm{sym}$}.
Then, $\mathbf{U}$ considers the global topology information of each node, which is in effect the same as considering the distant neighborhood in the graph.
As a result, \method performs well in different real-world datasets without tuning any hyperparameters in $\mathcal{P}$, unlike existing GNNs that are often required to increase the value of $K$ up to hundreds \cite{Wang21DGC}.

\paragraph{\underline{\smash{Time complexity}}}
The time complexity of \method, including all its components SVD, PCA, and the training of LR, is linear with a graph size in most real-world graphs where the number of edges is much larger than the numbers of nodes and features.

\begin{lemma}
    Given a graph, let $n$ and $e$ be the numbers of nodes and edges, respectively, and $d$ be the number of features. Then, the time complexity of the training of \method is $O(de + d^2n + d^3)$.
\label{lemma:complexity}
\end{lemma}

\begin{proof}
\spaceAroundProof
The proof is given in Appendix \ref{appendix:proof-complexity}.
\spaceAroundProof
\end{proof}

%% file: 050sanity.tex
\section{Proposed Sanity Checks}
\label{sec:unit-tests}

We propose a set of \emph{sanity checks} to directly evaluate the robustness of GNNs to various scenarios of node classification.

\subsection{Design of Sanity Checks}
\label{ssec:sanity-design}

We categorize possible scenarios of node classification based on the characteristics of node features $\mathbf{X}$, a graph structure $\mathbf{A}$, and node labels $\mathbf{y}$.
We denote by $A_{ij}$ and $Y_i$ the random variables for edge $(i, j)$ between nodes $i$ and $j$ and label $y_i$ of node $i$, respectively.
We summarize the nine possible cases in Figure \ref{fig:sanity-matrix}.

\header{Structure}
We consider three cases of the structure $\mathbf{A}$: \emph{uniform}, \emph{homophily}, and \emph{heterophily}, which are defined as follows:
\begin{compactitem}
	\item \textbf{Uniform:} $P(Y_i = y \mid A_{ij} = 1, Y_j = y) = P(Y_i = y)$
	\item \textbf{Homophily:} $P(Y_i = y \mid A_{ij} = 1, Y_j = y) > P(Y_i = y)$
	\item \textbf{Heterophily:} $P(Y_i = y \mid A_{ij} = 1, Y_j = y) < P(Y_i = y)$
\end{compactitem}

The uniform case means that the label of a node is independent of the labels of its neighbors.
This is the case when the graph structure provides no information for classification. 
In the homophily case, adjacent nodes are likely to have the same label, which is the most common assumption in graph data.
In the heterophily case, adjacent nodes are likely to have different labels, which is not as common as homophily but often observed in real-world graphs.

We assume the one-to-one correspondence between the structural property and the labels: \emph{uniform} with uniformly random $\mathbf{A}$, \emph{homophily} with block-diagonal $\mathbf{A}$, and \emph{heterophily} with non-block-diagonal $\mathbf{A}$.
Note that the other combinations, e.g., homophily with non-block-diagonal $\mathbf{A}$, or uniform with block-diagonal $\mathbf{A}$, are not feasible by definition.
Figure \ref{fig:sanity-all} illustrates the three cases of $\mathbf{A}$.
The number of node clusters in the graph, which is four in the figure, is the same as the number of node labels in experiments.

\header{Features}
We consider three cases of node features $\mathbf{X}$: \emph{random}, \emph{semantic}, and \emph{structural}.
The three cases are defined in relation to $\mathbf{A}$ and $\mathbf{y}$.
We use the notation $p(\cdot)$ since the features are typically modeled as continuous variables:
\begin{compactitem}
	\item \textbf{Random:} $p(\mathbf{x}_i, \mathbf{x}_j \mid y_i, y_j, a_{ij}) = p(\mathbf{x}_i, \mathbf{x}_j)$
	\item \textbf{Structural:} $p(\mathbf{x}_i, \mathbf{x}_j \mid y_i, y_j, a_{ij}) \neq p(\mathbf{x}_i, \mathbf{x}_j \mid y_i, y_j)$
	\item \textbf{Semantic:} $p(\mathbf{x}_i, \mathbf{x}_j \mid y_i, y_j, a_{ij}) \neq p(\mathbf{x}_i, \mathbf{x}_j \mid a_{ij})$
\end{compactitem}

The random case means that each feature element $x_{ij}$ is determined independently of all other variables in the graph, providing no useful information.
The semantic case represents a typical graph where $\mathbf{X}$ provides useful information of $\mathbf{y}$.
In this case, the feature $\mathbf{x}_i$ of each node $i$ is directly correlated with the label $y_i$.
In the structural case, $\mathbf{X}$ provides information of the graph structure, rather than the labels.
Thus, $\mathbf{X}$ is meaningful for classification if $\mathbf{A}$ is not \emph{uniform}, as it gives only indirect information of $\mathbf{y}$.

\subsection{Observations from Sanity Checks}

Table \ref{table:sanity-tests} shows the results of sanity checks for our \method and all baseline models whose details are given in Section \ref{ssec:exp-setup}.
We assume 4 target classes of nodes, making the accuracy of random guessing 25\%.
Among the nine cases in Table \ref{fig:sanity-matrix}, we do not report the results on two cases where the graph does not give any information (i.e., ``none helps''), since all methods produce similar accuracy.

\method wins on all sanity checks, thanks to its careful design for the robustness to various graph scenarios.
Although homophily $\mathbf{A}$ and useful (i.e., not random) $\mathbf{X}$ is a common assumption in many datasets, many nonlinear GNNs show failure (i.e., red cells) in such cases.
This implies that the \emph{theoretical expressiveness} of a model is often not aligned with its actual performance even in controlled testbeds, as we also show in our intuitive example in Figure~\ref{fig:example}.
Only a few baselines succeed in other scenarios with different $\mathbf{A}$ and $\mathbf{X}$, as we present in the observations below.

\begin{observation}[Summary]
\method wins on all sanity checks without failures (i.e., no red cells).
\method shows the best average accuracy and the average rank compared to 15 competitors.
\end{observation}

\begin{observation}[No network effects]
Only a few models including S$^2$GC and GraphSAGE perform well in uniform $\mathbf{A}$, where the graph structure provides no useful information, since they have the raw $\mathbf{X}$ (not multiplied with $\mathbf{A}$) in their propagator functions $\mathcal{P}$.
\end{observation}

\begin{observation}[Useless features]
None of the existing models in Table \ref{table:sanity-tests} succeeds with useless (i.e., random) features $\mathbf{X}$, since their propagator functions $\mathcal{P}$ rely on $\mathbf{X}$ in all cases.
\end{observation}

\begin{observation}[Heterophily graphs]
Models that can utilize even-hop neighbors, such as G$^2$CN, GraphSAGE, and GPR-GNN, succeed in heterophily graphs either with semantic or structural $\mathbf{X}$.
\end{observation}


\begin{table*}
\renewcommand{\arraystretch}{1.1}
\caption{
    \myTag{\method wins} most of the times on 13 real-world datasets ($7$ homophily and $6$ heterophily graphs) against 15 competitors.
    We color the best and worst results as green and red, respectively, as in Table~\ref{table:sanity-tests}.
    \method is the only approach that exhibits no failures (i.e., no red cells) in all datasets.
    Most competitors cause out-of-memory (OOM) errors on large graphs.
}
\spaceBelowTableCaption
\centering{\resizebox{\textwidth}{!}{
\begin{tabular}{ l | ccccccc | cccccc | r }
	\toprule
	\textbf{Model}
	    & \textbf{Cora}
	    & \textbf{CiteSeer} 
	    & \textbf{PubMed} 
	    & \textbf{Comp.} 
	    & \textbf{Photo} 
	    & \textbf{ArXiv}
	    & \textbf{Products}
	    & \textbf{Cham.} 
	    & \textbf{Squirrel} 
	    & \textbf{Actor} 
	    & \textbf{Penn94} 
	    & \textbf{Twitch}
	    & \textbf{Pokec}
        & \textbf{Avg. Rank} \\
	\midrule
	    LR
	    & \lose{51.5$\pm$1.2}
	    & \lose{52.9$\pm$4.5}
	    & \lose{79.9$\pm$0.5}
	    & \lose{73.9$\pm$1.2}
	    & \lose{79.3$\pm$1.5}
	    & \lose{48.3$\pm$1.9}
	    & \lose{56.4$\pm$0.5}
	    & \lose{24.9$\pm$1.7}
	    & \lose{26.7$\pm$1.9}
	    & 27.8$\pm$0.8
	    & 63.5$\pm$0.5
	    & \lose{53.0$\pm$0.1}
	    & \lose{61.3$\pm$0.0}
        & 11.7 (4.2) \\
	\midrule
	Reg. Kernel
	    & \lose{67.8$\pm$2.5}
	    & 62.1$\pm$4.4
	    & \lose{83.4$\pm$1.4}
	    & \lose{80.3$\pm$1.4}
	    & \lose{87.1$\pm$1.2}
	    & \oom{O.O.M.}
	    & \oom{O.O.M.}
	    & \lose{29.4$\pm$2.6}
	    & \lose{24.3$\pm$2.3}
	    & \silver{29.6$\pm$1.4}
	    & \oom{O.O.M.}
	    & \oom{O.O.M.}
	    & \oom{O.O.M.}
        & 12.2 (3.8) \\
	Diff. Kernel
	    & \lose{70.6$\pm$1.5}
	    & 62.7$\pm$3.8
	    & \lose{82.1$\pm$0.4}
	    & \lose{83.1$\pm$1.0}
	    & \lose{89.8$\pm$0.6}
	    & \oom{O.O.M.}
	    & \oom{O.O.M.}
	    & 34.5$\pm$7.9
	    & 28.3$\pm$1.5
	    & \lose{24.7$\pm$0.9}
	    & \lose{53.5$\pm$0.8}
	    & \oom{O.O.M.}
	    & \oom{O.O.M.}
        & 11.8 (2.5) \\
	RW Kernel
	    & \lose{72.7$\pm$1.7}
	    & 64.1$\pm$3.9
	    & \lose{83.1$\pm$0.7}
	    & 84.2$\pm$0.7
	    & \lose{90.6$\pm$0.7}
	    & 63.2$\pm$0.2
	    & \lose{74.2$\pm$0.0}
	    & 34.9$\pm$3.5
	    & \lose{25.0$\pm$1.6}
	    & \lose{26.4$\pm$1.1}
	    & 63.1$\pm$0.7
	    & \lose{57.6$\pm$0.1}
	    & \lose{59.5$\pm$0.0}
        & 8.3 (3.3) \\
	\midrule
	SGC
	    & 76.2$\pm$1.1
	    & 65.8$\pm$3.9
	    & 84.1$\pm$0.8
	    & 83.7$\pm$1.6
	    & \lose{90.1$\pm$0.9}
	    & \bronze{65.0$\pm$3.4}
	    & \lose{74.6$\pm$5.1}
	    & 38.1$\pm$4.5
	    & \gold{33.1$\pm$1.0}
	    & \lose{24.6$\pm$0.8}
	    & \bronze{64.0$\pm$1.1}
	    & \lose{56.5$\pm$0.1}
	    & \silver{69.8$\pm$0.0}
        & 6.6 (4.2) \\
	DGC
	    & 77.8$\pm$1.4
	    & \bronze{66.1$\pm$4.2}
	    & 84.3$\pm$0.6
	    & 83.9$\pm$0.7
	    & \lose{90.4$\pm$0.2}
	    & \silver{65.2$\pm$4.0}
	    & \lose{68.7$\pm$13.}
	    & 37.2$\pm$3.7
	    & 29.2$\pm$1.2
	    & \lose{25.2$\pm$2.1}
	    & 62.5$\pm$0.4 
	    & \bronze{58.2$\pm$0.2}
	    & \lose{60.7$\pm$0.1}
        & 6.6 (3.2) \\
	S$^{2}$GC
	    & \bronze{78.3$\pm$1.5}
	    & \silver{66.9$\pm$4.4}
	    & 84.3$\pm$0.3
	    & \lose{83.1$\pm$0.8}
	    & \lose{90.1$\pm$0.8}
	    & 62.0$\pm$7.4
	    & \lose{58.3$\pm$18.}
	    & 34.9$\pm$4.9
	    & \lose{27.6$\pm$1.8}
	    & \lose{26.7$\pm$1.8}
	    & 63.1$\pm$0.5 
	    & \silver{58.7$\pm$0.1}
	    & \lose{61.2$\pm$0.0}
        & 6.6 (2.7) \\
	G$^{2}$CN
	    & 76.6$\pm$1.5
	    & 64.2$\pm$3.3
	    & \lose{81.4$\pm$0.6}
	    & \lose{82.8$\pm$1.6}
	    & \lose{88.8$\pm$0.5}
	    & \oom{O.O.M.}
	    & \oom{O.O.M.}
	    & \silver{40.7$\pm$2.9}
	    & \silver{32.1$\pm$1.5}
	    & \lose{24.3$\pm$0.5}
	    & \oom{O.O.M.}
	    & \oom{O.O.M.}
	    & \oom{O.O.M.}
        & 10.5 (4.5) \\
	\midrule
	GCN
	  & 76.0$\pm$1.2
	    & 65.0$\pm$2.9
	    & 84.3$\pm$0.5
	    & \bronze{85.1$\pm$0.9}
	    & 91.6$\pm$0.5
	    & 62.8$\pm$0.6
	    & \oom{O.O.M.}
	    & 38.5$\pm$3.0
	    & \bronze{31.4$\pm$1.8}
	    & \lose{26.8$\pm$0.4}
	    & 62.9$\pm$0.7 
	    & \lose{57.0$\pm$0.1}
	    & \lose{63.9$\pm$0.4}
        & \bronze{6.3 (2.4)} \\
	SAGE
	    & \lose{74.6$\pm$1.3}
	    & 63.7$\pm$3.6
	    & \lose{82.9$\pm$0.4}
	    & 83.8$\pm$0.5
	    & \lose{90.6$\pm$0.5}
	    & 61.5$\pm$0.6
	    & \oom{O.O.M.}
	    & \bronze{39.8$\pm$4.3}
	    & \lose{27.0$\pm$1.3}
	    & 27.8$\pm$0.9
	    & \oom{O.O.M.} 
	    & \lose{56.6$\pm$0.4}
	    & \bronze{68.9$\pm$0.1}
        & 8.5 (3.5) \\
	GCNII
	    & 77.8$\pm$1.7
	    & 63.4$\pm$3.0
	    & \bronze{84.9$\pm$0.8}
	    & \lose{82.3$\pm$1.8}
	    & 90.8$\pm$0.6
	    & \lose{45.7$\pm$0.5}
	    & \oom{O.O.M.}
	    & \lose{30.5$\pm$2.5}
	    & \lose{21.9$\pm$3.0}
	    & 29.0$\pm$1.3
	    & \silver{64.5$\pm$0.5} 
	    & \lose{56.9$\pm$0.6}
	    & \lose{62.1$\pm$0.3}
        & 8.4 (4.6) \\
    H$^{2}$GCN
	    & 77.6$\pm$0.9
	    & 64.7$\pm$3.8
	    & \gold{85.4$\pm$0.4}
	    & \lose{49.5$\pm$16.}
	    & \lose{75.8$\pm$11.}
	    & \oom{O.O.M.}
	    & \oom{O.O.M.}
	    & 31.9$\pm$2.6
	    & \lose{25.0$\pm$0.5}
	    & 28.9$\pm$0.6
	    & 63.9$\pm$0.4
	    & \silver{58.7$\pm$0.0}
	    & \oom{O.O.M.}
        & 8.9 (4.9) \\
	APPNP
	    & \gold{80.0$\pm$0.6}
	    & \gold{67.1$\pm$2.8}
	    & 84.6$\pm$0.5
	    & 84.2$\pm$1.7
	    & \silver{92.5$\pm$0.3}
	    & \lose{53.4$\pm$1.3}
	    & \oom{O.O.M.}
	    & \lose{30.9$\pm$4.7}
	    & \lose{23.9$\pm$3.2}
	    & \lose{26.1$\pm$1.0}
	    & 63.7$\pm$0.9 
	    & \lose{47.3$\pm$0.3}
	    & \lose{57.4$\pm$0.4}
        & 7.6 (4.8) \\
	GPR-GNN
	    & \silver{78.8$\pm$1.3}
	    & 64.2$\pm$4.0
	    & \silver{85.1$\pm$0.7}
	    & 85.0$\pm$1.0
	    & \gold{92.6$\pm$0.3}
	    & 58.5$\pm$0.8
	    & \oom{O.O.M.}
	    & 31.7$\pm$4.7
	    & \lose{26.2$\pm$1.6}
	    & \bronze{29.5$\pm$1.1}
	    & \silver{64.5$\pm$0.4} 
	    & \lose{57.6$\pm$0.2}
	    & \lose{67.6$\pm$0.1}
        & \silver{5.4 (3.7)} \\
	GAT
	    & 78.2$\pm$1.2
	    & 65.8$\pm$4.0
	    & 83.6$\pm$0.2
	    & \silver{85.4$\pm$1.4}
	    & 91.7$\pm$0.5
	    & 58.2$\pm$1.0
	    & \oom{O.O.M.}
	    & 39.1$\pm$4.1
	    & 28.6$\pm$0.6
	    & \lose{26.4$\pm$0.4}
	    & \lose{60.5$\pm$0.8} 
	    & \oom{O.O.M.}
	    & \oom{O.O.M.}
        & 7.5 (3.7) \\
	\midrule
	\method
	    & 77.8$\pm$1.1
	    & \gold{67.1$\pm$2.3}
	    & 84.6$\pm$0.5
	    & \gold{86.3$\pm$0.7}
	    & \bronze{91.8$\pm$0.5}
	    & \gold{66.3$\pm$0.3}
	    & \gold{84.9$\pm$0.0} 
	    & \gold{40.8$\pm$3.2}
	    & 31.1$\pm$0.7
	    & \gold{30.9$\pm$0.6}
	    & \gold{68.2$\pm$0.6} 
	    & \gold{59.7$\pm$0.1}
	    & \gold{73.9$\pm$0.1}
        & \gold{1.9 (1.5)} \\
	\bottomrule
\end{tabular}
}}
\label{table:expr}
\spaceBelowLargeTable
\end{table*}



\begin{table}
\centering
\caption{
    \myTag{Dataset statistics.}
    The first $7$ datasets are homophily, and the last $6$ are heterophily graphs.
    }
\spaceBelowTableCaption
\small
\begin{tabular}{l|rrrr}
    \toprule
    \textbf{Dataset} & \textbf{Nodes} & \textbf{Edges} & \textbf{Features} & \textbf{Classes} \\
    \midrule
    \textbf{Cora} & 2,708 & 5,429 & 1433 & 7 \\
    \textbf{CiteSeer} & 3,327 & 4,732 & 3703 & 6 \\
    \textbf{PubMed} & 19,717 & 44,338 & 500 & 3 \\
    \textbf{Computers} & 13,752 & 245,861 & 767 & 10 \\
    \textbf{Photo} & 7,650 & 119,081 & 745 & 8 \\
    \textbf{ogbn-arXiv} & 169,343 & 1,166,243 & 128 & 40 \\
    \textbf{ogbn-Products} & 2,449,029 & 61,859,140 & 100 & 30 \\
    \midrule
    \textbf{Chameleon} & 2,277 & 36,101 & 2325 & 5 \\
    \textbf{Squirrel} & 5,201 & 216,933 & 2089 & 5 \\
    \textbf{Actor} & 7,600 & 29,926 & 931 & 5 \\
    \textbf{Penn94} & 41,554 & 1,362,229 & 4814 & 2 \\
    \textbf{Twitch} & 168,114 & 6,797,557 & 7 & 2 \\
    \textbf{Pokec} & 1,632,803 & 30,622,564 & 65 & 2 \\
    \bottomrule
\end{tabular}
\label{table:datasets}
\spaceBelowSmallTable
\end{table}


%% file: 060exp.tex
\section{Experiments}
\label{sec:exp}

We conduct experiments on $13$ real-world datasets to answer the following research questions (RQ):
\begin{compactenum}[{RQ}1.]
    \item {\bf Accuracy:} How well does \method work for semi-supervised node classification on real-world graphs?
    \item {\bf Success of simplicity:} How does \method succeed even with its simplicity? What if we add nonlinearity to \method?
    \item {\bf Speed and scalability:} How fast and scalable is \method to large real-world graphs?
    \item {\bf Interpretability:} How to explain the importance of graph signals through the learned weights of \method?
    \item {\bf Ablation study:} Are all the design decisions of \method, such as two-hop aggregation, effective in real-world graphs?
\end{compactenum}

\subsection{Experimental Setup}
\label{ssec:exp-setup}

We introduce our experimental setup including datasets, evaluation processes, and baselines for node classification.

\header{Datasets}
We use $7$ homophily and $6$ heterophily graph datasets in experiments, which were commonly used in previous works on node classification \citep{Chien21GPRGNN, Pei20GeomGCN, Lim21LINKX}.
Table \ref{table:datasets} shows a summary of dataset information.
Cora, CiteSeer, and PubMed \citep{Sen08Data, Yang16revisiting} are homophily citation graphs between research articles.
Computers and Photo \citep{Shchur18pitfalls} are homophily Amazon co-purchase graphs between items.
ogbn-arXiv and ogbn-Products are large homophily graphs from Open Graph Benchmark \citep{hu2020ogb}.
Since we use only $2.5\%$ of all labels as training data, we omit the classes with instances fewer than $100$.
Chameleon and Squirrel \citep{Rozemberczki21data} are heterophily Wikipedia web graphs.
Actor \citep{Tang09data} is a heterophily graph connected by the co-occurrence of actors on Wikipedia pages.
Penn94 \citep{traud2012social, Lim21LINKX} is a heterophily graph of gender relations in a social network.
Twitch \citep{rozemberczki2021twitch} and Pokec \citep{leskovec2014snap} are large graphs, which have been relabeled by \citep{Lim21LINKX} to be heterophily.
We make the heterophily graphs undirected as done in \citep{Chien21GPRGNN}.


\begin{table*}
\caption{
	\myTag{\method effectively combines different components.}
    \method-C$i$ represents that we use only the $i$-th component shown in \Eqref{eq:proposed}.
    Node classification is done accurately even we use a single component at each time, and \method outperforms the best accuracy of a single component in 11 out of the 13 datasets.
    Green (\colorbox{green!50}{}, \colorbox{green!20}{}) marks the top two.
}
\spaceBelowTableCaption
\centering{\resizebox{\textwidth}{!}{
\begin{tabular}{ l | ccccccc | cccccc }
	\toprule
	\textbf{Model} 
	    & \textbf{Cora}
	    & \textbf{CiteSeer}
	    & \textbf{PubMed} 
	    & \textbf{Comp.}
	    & \textbf{Photo} 
	    & \textbf{ArXiv} 
	    & \textbf{Products}
	    & \textbf{Cham.} 
	    & \textbf{Squirrel} 
	    & \textbf{Actor} 
	    & \textbf{Penn94} 
	    & \textbf{Twitch}
	    & \textbf{Pokec} \\
	\midrule
    \method-C$1$
	    & 46.3$\pm$3.0
	    & 29.2$\pm$2.5
	    & 64.5$\pm$1.0
	    & 77.6$\pm$1.0
	    & 78.5$\pm$0.9
	    & 51.4$\pm$0.2
	    & 73.6$\pm$2.9
	    & \goldtwo{41.9$\pm$2.0}
	    & 29.1$\pm$1.0
	    & 21.6$\pm$1.2
	    & 60.9$\pm$0.6
	    & \silvertwo{59.3$\pm$0.1}
	    & 66.7$\pm$0.0 \\
    \method-C$2$
	    & 53.5$\pm$1.5
	    & 53.6$\pm$3.6
	    & 79.3$\pm$0.3
	    & 74.5$\pm$1.1
	    & 81.4$\pm$0.9
	    & 49.8$\pm$0.2
	    & 57.4$\pm$0.1
	    & 25.1$\pm$1.5
	    & 21.8$\pm$0.9
	    & \silvertwo{29.9$\pm$2.1}
	    & 62.3$\pm$0.5
	    & 53.0$\pm$0.1
	    & 61.1$\pm$0.0 \\
    \method-C$3$
	    & \silvertwo{77.6$\pm$0.7}
	    & 62.7$\pm$4.3
	    & 77.4$\pm$0.8
	    & \silvertwo{86.0$\pm$1.0}
	    & 90.3$\pm$0.8
	    & \silvertwo{66.2$\pm$0.2}
	    & \silvertwo{82.5$\pm$0.0}
	    & 40.6$\pm$1.0
	    & 27.6$\pm$3.2
	    & 24.1$\pm$1.4
	    & \silvertwo{64.7$\pm$0.6}
	    & 53.1$\pm$0.1
	    & \silvertwo{73.2$\pm$0.1} \\
    \method-C$4$
	    & 76.8$\pm$0.9
	    & \silvertwo{64.7$\pm$3.6}
	    & \silvertwo{82.1$\pm$0.7}
	    & 85.3$\pm$1.2
	    & \silvertwo{90.9$\pm$0.7}
	    & 65.3$\pm$0.3
	    & 78.3$\pm$0.1
	    & 40.4$\pm$2.2
	    & \goldtwo{31.6$\pm$1.4}
	    & 23.7$\pm$1.4
	    & 64.3$\pm$0.7
	    & 56.3$\pm$0.1
	    & 68.4$\pm$0.2  \\
    \midrule
    \method
	    & \goldtwo{77.8$\pm$1.1}
	    & \goldtwo{67.1$\pm$2.3}
	    & \goldtwo{84.6$\pm$0.5}
	    & \goldtwo{86.3$\pm$0.7}
	    & \goldtwo{91.8$\pm$0.5}
	    & \goldtwo{66.3$\pm$0.3}
	    & \goldtwo{84.9$\pm$0.0}
	    & \silvertwo{40.8$\pm$3.2}
	    & \silvertwo{31.1$\pm$0.7}
	    & \goldtwo{30.9$\pm$0.6}
	    & \goldtwo{68.2$\pm$0.6}
	    & \goldtwo{59.7$\pm$0.1}
	    & \goldtwo{73.9$\pm$0.1} \\
	\bottomrule
\end{tabular}}}
\label{table:ablation-components}
\spaceBelowLargeTable
\end{table*}



\begin{table*}
\centering
\caption{
	\myTag{Linearity is enough for \method:} it outperforms its own variants with nonlinearity that replace the linear classifier or the PCA function $g$ with a nonlinear neural network.
    Green (\colorbox{green!50}{}, \colorbox{green!20}{}) marks the top two.
}
\spaceBelowTableCaption
\resizebox{\textwidth}{!}{
\begin{tabular}{ l | ccccccc | cccccc }
	\toprule
	\textbf{Model} 
	    & \textbf{Cora} 
	    & \textbf{CiteSeer} 
	    & \textbf{PubMed} 
	    & \textbf{Comp.} 
	    & \textbf{Photo} 
	    & \textbf{ArXiv} 
	    & \textbf{Products} 
	    & \textbf{Cham.} 
	    & \textbf{Squirrel} 
	    & \textbf{Actor} 
	    & \textbf{Penn94} 
	    & \textbf{Twitch} 
	    & \textbf{Pokec} \\
	\midrule
	w/ MLP-2
	    & 65.9$\pm$1.1
	    & 54.2$\pm$5.3
	    & \silvertwo{83.3$\pm$0.2}
	    & 84.8$\pm$0.5
	    & \silvertwo{90.1$\pm$1.9}
	    & \silvertwo{65.8$\pm$0.1}
	    & \goldtwo{85.7$\pm$0.0}
	    & 40.0$\pm$2.5
	    & 30.5$\pm$0.5
	    & 28.8$\pm$1.0
	    & \silvertwo{66.7$\pm$1.7}
	    & \silvertwo{60.3$\pm$0.3}
	    & \goldtwo{76.5$\pm$0.1} \\
	w/ MLP-3
	    & 66.1$\pm$2.0
	    & 50.3$\pm$3.6
	    & 80.9$\pm$0.9
	    & \silvertwo{85.2$\pm$0.8}
	    & 90.0$\pm$0.8
	    & 63.0$\pm$0.1
	    & \silvertwo{84.9$\pm$0.9}
	    & 38.5$\pm$5.5
	    & \silvertwo{30.9$\pm$0.7}
	    & \silvertwo{28.9$\pm$1.2}
	    & 65.1$\pm$0.6
	    & \silvertwo{60.3$\pm$0.2}
	    & \silvertwo{76.4$\pm$0.2} \\
	w/ NL Trans.
	    & \silvertwo{70.7$\pm$2.3}
	    & \silvertwo{57.5$\pm$5.1}
	    & 81.0$\pm$0.4
	    & 71.4$\pm$10. 
	    & 77.9$\pm$2.2
	    & 57.0$\pm$0.6
	    & O.O.M.
	    & \goldtwo{41.3$\pm$3.2}
	    & 30.0$\pm$1.6
	    & 27.6$\pm$2.4
	    & 61.8$\pm$1.6
	    & \goldtwo{61.5$\pm$0.3}
	    & 75.7$\pm$0.5 \\
    \midrule
    \method
	    & \goldtwo{77.8$\pm$1.1}
	    & \goldtwo{67.1$\pm$2.3}
	    & \goldtwo{84.6$\pm$0.5}
	    & \goldtwo{86.3$\pm$0.7}
	    & \goldtwo{91.8$\pm$0.5}
	    & \goldtwo{66.3$\pm$0.3}
	    & \silvertwo{84.9$\pm$0.0}
	    & \silvertwo{40.8$\pm$3.2}
	    & \goldtwo{31.1$\pm$0.7}
	    & \goldtwo{30.9$\pm$0.6}
	    & \goldtwo{68.2$\pm$0.6}
	    & 59.7$\pm$0.1
	    & 73.9$\pm$0.1 \\
	\bottomrule
\end{tabular}}
\label{table:ablation-linearity}
\spaceBelowLargeTable
\end{table*}


\header{Evaluation}
We perform semi-supervised node classification by randomly dividing all nodes in a graph by the $2.5\%/2.5\%/95\%$ ratio into training, validation, and test sets.
In this setting \cite{Chien21GPRGNN, Ding22Meta}, which is common in real-world data where labels are often scarce and expensive, we can properly evaluate the performance of each method in semi-supervised learning.
We perform five runs of each experiment with different random seeds and report the average and standard deviation.
All hyperparameter searches and early stopping are done based on validation accuracy for each run.

\header{Competitors and hyperparameters}
We include various types of competitors: linear models (LR, SGC, DGC, S$^2$GC, and G$^2$CN), coupled nonlinear models (GCN, GraphSAGE, GCNII, and H$_2$GCN), decoupled models (APPNP and GPR-GNN), and attention-based models (GAT).
We perform row-normalization on the node features as done in most studies on GNNs.
We search their hyperparameters for every data split through a grid search as described in Appendix \ref{appendix:reproducibility}.
The hidden dimension size is set to $64$, and the dropout probability is set to $0.5$.
For the linear models, we use L-BFGS to train $100$ epochs with the patience of $5$.
For the nonlinear ones, we use ADAM and update them for $1000$ epochs with the patience of $200$.

We also include three graph kernel methods \citep{Smola03Kernels}, namely Regularized Laplacian, Diffusion Process, and the $K$-step Random Walk.
They focus on feature transformation and are used with the LR classifier as \method is.
Thus, they perform as the direct competitors of \method that apply different feature transformations.
The propagator functions of graph kernel methods \citep{Smola03Kernels} are given as follows:
\begin{align}
    & \textrm{(Reg. Kernel)} \ \mathcal{P}(\mathbf{A}, \mathbf{X}) = (\mathbf{I}_n + \sigma^2 \tilde{\mathbf{L}})^{-1} \mathbf{X} \\
    & \textrm{(Diff. Kernel)} \ \mathcal{P}(\mathbf{A}, \mathbf{X}) = \mathrm{exp}(-\sigma^2 / 2\tilde{\mathbf{L}}) \mathbf{X} \\
    & \textrm{(RW Kernel)} \ \mathcal{P}(\mathbf{A}, \mathbf{X}) = (a \mathbf{I}_n - \tilde{\mathbf{L}})^p \mathbf{X},
\end{align}
where \smash{$\tilde{\mathbf{L}} = \mathbf{D}^{-1/2} (\mathbf{D} - \mathbf{A}) \mathbf{D}^{-1/2}$} is the normalized Laplacian matrix, and $\sigma=1$, $a=1$, and $p=2$ are their hyperparameters.

\method contains two hyperparameters, which are the weight of LASSO and the weight of group LASSO.
It is worth noting that \method does not need to recompute the features when searching the hyperparameters, while most of the linear methods need to do so because of including one or more hyperparameters in $\mathcal{P}$.

\subsection{Accuracy (RQ1)}

In Table~\ref{table:expr}, \method is compared against $15$ competitors on $13$ real-world datasets ($7$ homophily and $6$ heterophily graphs).
We color the best and worst results as green and red, respectively, as in Table \ref{table:sanity-tests}.
We report the accuracy in Table~\ref{table:expr} where \colorbox{green}{}, \colorbox{green!40}{}, \colorbox{green!20}{} represent the top three methods (higher is darker),
\method outperforms all competitors in 4 homophily and 5 heterophily graphs, and achieves competitive accuracy in the rest; \method is among the top three in $10$ out of $13$ times.
Moreover, \method is the only model that exhibits no failures (i.e., no red cells), and shows the best average rank with a significant difference from the second-best.
It is notable that many competitors, even linear models such as the kernel methods and G$^2$CN, run out of memory in large graphs.
This shows that linearity is not a sufficient condition for efficiency and scalability, and thus a careful design of the propagator function $\mathcal{P}$ is needed as in \method.

\vspace{-1mm}
\subsection{Success of Simplicity (RQ2)}
We conduct studies to better understand how \method exhibits the superior performance on real-world graphs even with its simplicity.
Tables \ref{table:ablation-components} and \ref{table:ablation-linearity} demonstrate the strength of simplicity for semi-supervised node classification in real-world graphs.

\header{Flexibility}
Table \ref{table:ablation-components} illustrates the accuracy of \method when only one of its four components in \Eqref{eq:proposed} is used at each time.
The accuracy of \method with only a single component is higher than those of most baselines in Table \ref{table:expr} when an appropriate component is picked for each dataset, e.g., C1 for Twitch, C2 for Actor, C3 for Cora, and C4 for CiteSeer.
This shows that high accuracy in semi-supervised node classification can be achieved by a well-designed simple model even without high \emph{expressivity}.
\method focuses on the best component in each dataset effectively, improving the accuracy of individual components in 11 out of 13 datasets.

\header{Nonlinearity}
Many recent works on linear GNNs have shown that nonlinearity is not an essential component in semi-supervised node classification \cite{Zhu21S2GC, Wang21DGC, Li22G2CN}.
To support the success of \method, we design three nonlinear variants of it:
\begin{compactitem}
	\item \textbf{w/ MLP-2:} We replace LR with a 2-layer MLP.
	\item \textbf{w/ MLP-3:} We replace LR with a 3-layer MLP.
	\item \textbf{w/ Nonlinear (NL) Transformation:}
	    We replace the PCA function $g(\cdot)$ as a nonlinear function. 
	    Specifically, we adopt a 2-layer MLP for the first two components and a 2-layer GCN for the last two components.
	    The transformed features are concatenated and given to another 2-layer MLP.
\end{compactitem}


\begin{table}
\caption{
	\myTag{\method is scalable.} 
    We measure the runtime of \method by varying the numbers of features and edges in a graph.
}
\small
\spaceBelowTableCaption
\resizebox{0.74\columnwidth}{!}{
\begin{tabular}{ l | ccccc }
	\toprule
	\textbf{Number of features} & 20 & 40 & 60 & 80 & 100 \\
	\midrule
	\textbf{Time (s)} & 13.0 & 16.2 & 18.6 & 22.9 & 27.8 \\
	\midrule
	\midrule
	\textbf{Number of edges} & 12M & 25M & 60M & 80M & 100M \\
	\midrule
	\textbf{Time (s)} & 12.3 & 19.9 & 21.9 & 24.0 & 27.8 \\
    \bottomrule
\end{tabular}
}
\label{table:scalability}
\spaceBelowSmallTable
\end{table}


We use dropout with a probability of $0.5$ to prevent overfitting in both MLP and GCN.
The nonlinear models are trained with the same setting as GCN reported in Table~\ref{table:expr}.
We report the result in Table~\ref{table:ablnl}, showing that adding nonlinearity does not necessarily improve the accuracy while sacrificing both scalability and interpretability.


\begin{table*}
\centering
\caption{
	\myTag{Ablation study - \method works best with the current design.}
    Each design decision of \method leads to an improvement of accuracy in real-world graphs; \method is among the top two in all datasets, marked as green (\colorbox{green!50}{}, \colorbox{green!20}{}).
}
\spaceBelowTableCaption
\resizebox{\textwidth}{!}{
\begin{tabular}{ l | ccccccc | cccccc }
	\toprule
	\textbf{Model} 
	    & \textbf{Cora} 
	    & \textbf{CiteSeer} 
	    & \textbf{PubMed} 
	    & \textbf{Comp.} 
	    & \textbf{Photo} 
	    & \textbf{ArXiv} 
	    & \textbf{Products} 
	    & \textbf{Cham.} 
	    & \textbf{Squirrel} 
	    & \textbf{Actor} 
	    & \textbf{Penn94} 
	    & \textbf{Twitch} 
	    & \textbf{Pokec} \\
	\midrule
	w/o Sp. Reg.
	    & \silvertwo{77.8$\pm$0.6}
	    & 65.0$\pm$3.5
	    & 83.8$\pm$0.5
	    & 85.9$\pm$0.8
	    & 91.7$\pm$0.7
	    & 65.2$\pm$0.2
	    & 83.4$\pm$1.9
	    & 40.1$\pm$3.8
	    & 30.7$\pm$1.0
	    & 30.1$\pm$0.6
	    & 67.4$\pm$0.6
	    & \goldtwo{59.8$\pm$0.1}
	    & \goldtwo{74.2$\pm$0.0} \\
	w/o PCA
	    & 74.8$\pm$1.5
	    & 66.0$\pm$3.1
	    & \goldtwo{84.7$\pm$0.5}
	    & 84.4$\pm$1.1
	    & 90.3$\pm$0.7
	    & 60.8$\pm$0.2
	    & \silvertwo{84.5$\pm$0.0}
	    & \goldtwo{41.3$\pm$2.0}
	    & \goldtwo{31.8$\pm$1.1}
	    & 27.3$\pm$1.1
	    & \silvertwo{67.7$\pm$0.7}
	    & 59.1$\pm$0.2
	    & 72.8$\pm$0.1 \\
	w/o Struct. $\mathbf{U}$
	    & \goldtwo{78.1$\pm$1.0}
	    & \goldtwo{67.4$\pm$2.5}
	    & 84.4$\pm$0.3
	    & \silvertwo{86.0$\pm$0.5}
	    & \goldtwo{92.1$\pm$0.4}
	    & \silvertwo{66.1$\pm$0.2}
	    & 82.5$\pm$0.6
	    & 37.5$\pm$4.2
	    & 29.8$\pm$0.4
	    & \goldtwo{31.3$\pm$0.5}
	    & 65.8$\pm$0.6
	    & 56.8$\pm$0.0
	    & 72.8$\pm$0.1 \\
    \midrule
    \method
	    & \silvertwo{77.8$\pm$1.1}
	    & \silvertwo{67.1$\pm$2.3}
	    & \silvertwo{84.6$\pm$0.5}
	    & \goldtwo{86.3$\pm$0.7}
	    & \silvertwo{91.8$\pm$0.5}
	    & \goldtwo{66.3$\pm$0.3}
	    & \goldtwo{84.9$\pm$0.0}
	    & \silvertwo{40.8$\pm$3.2}
	    & \silvertwo{31.1$\pm$0.7}
	    & \silvertwo{30.9$\pm$0.6}
	    & \goldtwo{68.2$\pm$0.6}
	    & \silvertwo{59.7$\pm$0.1}
	    & \silvertwo{73.9$\pm$0.1} \\
	\bottomrule
\end{tabular}}
\label{table:ablnl}
\spaceBelowLargeTable
\end{table*}



\begin{table*}
\caption{
	\myTag{Ablation study - Two-step aggregation is good enough for \method.} 
    The values $k_\mathrm{row}$ and $k_\mathrm{sym}$ represent the numbers of propagation steps for the $\mathbf{A}_\mathrm{row}$ and \smash{$\tilde{\mathbf{A}}_\mathrm{sym}$} components in \Eqref{eq:proposed}, respectively.
    We observe no consistent improvement of accuracy by increasing the values of $k_\mathrm{row}$ and $k_\mathrm{sym}$, supporting the current design of \method.
	Green (\colorbox{green!50}{}, \colorbox{green!20}{}) marks the top two.
}
\spaceBelowTableCaption
\centering{\resizebox{\textwidth}{!}{
\begin{tabular}{ ll | ccccccc | cccccc }
	\toprule
	\textbf{$k_{\text{row}}$} 
	    & \textbf{$k_{\text{sym}}$}
	    & \textbf{Cora}
	    & \textbf{CiteSeer}
	    & \textbf{PubMed} 
	    & \textbf{Comp.}
	    & \textbf{Photo} 
	    & \textbf{ArXiv} 
	    & \textbf{Products}
	    & \textbf{Cham.} 
	    & \textbf{Squirrel} 
	    & \textbf{Actor} 
	    & \textbf{Penn94} 
	    & \textbf{Twitch}
	    & \textbf{Pokec} \\
	\midrule
	$\{2, 4, 6\}$ & $\{2, 3, 4\}$
	    & \goldtwo{79.4$\pm$1.1}
	    & 66.0$\pm$4.4
	    & \silvertwo{84.4$\pm$0.4}
	    & 85.9$\pm$0.5
	    & 91.3$\pm$0.5
	    & \goldtwo{67.9$\pm$0.2}
	    & 84.8$\pm$2.1
	    & \silvertwo{41.0$\pm$3.9}
	    & \goldtwo{31.4$\pm$0.6}
	    & 29.8$\pm$0.6
	    & \silvertwo{68.2$\pm$0.6}
	    & \silvertwo{60.7$\pm$0.1}
	    & \goldtwo{76.6$\pm$0.2} \\
    $\{2, 4\}$ & $\{2, 3\}$
	    & 78.9$\pm$0.9
	    & \silvertwo{66.9$\pm$2.5}
	    & 84.0$\pm$0.5
	    & \silvertwo{86.2$\pm$0.7}
	    & \silvertwo{91.5$\pm$0.5}
	    & \silvertwo{67.4$\pm$0.1}
	    & 73.9$\pm$23.
	    & \goldtwo{41.4$\pm$4.0}
	    & 31.0$\pm$0.7
	    & 30.4$\pm$0.4
	    & \goldtwo{68.3$\pm$0.5}
	    & \goldtwo{60.8$\pm$0.1}
	    & \silvertwo{76.0$\pm$0.1} \\
    \midrule
	$6$ & $4$
	    & \silvertwo{79.2$\pm$0.7}
	    & 66.2$\pm$3.4
	    & 84.0$\pm$0.3
	    & 85.2$\pm$0.7
	    & 91.0$\pm$0.8
	    & 67.3$\pm$0.2
	    & 84.7$\pm$1.7
	    & 38.2$\pm$6.3
	    & 29.2$\pm$1.5
	    & \goldtwo{31.2$\pm$0.8}
	    & 67.7$\pm$0.7
	    & 59.8$\pm$0.1
	    & 74.2$\pm$0.2 \\
	$4$ & $3$
	    & \silvertwo{79.2$\pm$0.8}
	    & 66.1$\pm$3.5
	    & 84.2$\pm$0.5
	    & 85.8$\pm$0.6
	    & 91.3$\pm$0.4
	    & 67.2$\pm$0.1
	    & \goldtwo{85.4$\pm$0.0}
	    & 38.6$\pm$7.1
	    & 29.5$\pm$1.8
	    & 30.4$\pm$0.7
	    & 67.5$\pm$0.6
	    & 60.0$\pm$0.1
	    & 74.6$\pm$0.1 \\
	\midrule
	$2$ (ours) & $2$ (ours)
	    & 77.8$\pm$1.1
	    & \goldtwo{67.1$\pm$2.3}
	    & \goldtwo{84.6$\pm$0.5}
	    & \goldtwo{86.3$\pm$0.7}
	    & \goldtwo{91.8$\pm$0.5}
	    & 66.3$\pm$0.3
	    & \silvertwo{84.9$\pm$0.0}
	    & 40.8$\pm$3.2
	    & \silvertwo{31.1$\pm$0.7}
	    & \silvertwo{30.9$\pm$0.6}
	    & \silvertwo{68.2$\pm$0.6}
	    & 59.7$\pm$0.1
	    & 73.9$\pm$0.1 \\
	\bottomrule
\end{tabular}}}
\label{table:ablrf}
\spaceBelowLargeTable
\end{table*}


\subsection{Speed and Scalability (RQ3)}

We plot the training time versus the accuracy of each model on the ogbn-arXiv, ogbn-Products, and Pokec graphs, which are the largest in our benchmark, in Figure~\ref{fig:overview-0}.
We report the training time of each model with the hyperparameters that show the highest validation accuracy.
\method achieves the highest accuracy in the ogbn-arXiv and ogbn-Products datasets while being $10.4\times$ and $2.5\times$ faster than the second-best model, respectively.
\method also shows the highest accuracy in the Pokec dataset, while being $18.0\times$ faster than the best-performing deep model.
It is worth noting that \method is even faster than LR in ogbn-arXiv, taking fewer iterations than in LR during the optimization.
Its fast convergence is owing to the orthogonalization of each component of the features.

Table \ref{table:scalability} shows the training time of \method in the ogbn-Products graph with varying numbers of features and edges.
Smaller graphs are created by removing edges uniformly at random.
\method scales well with both variables even in large graphs containing up to 61M edges, showing linear complexity as we claim in Lemma \ref{lemma:complexity}.


\begin{figure}
    \centering
    \begin{subfigure}{0.19\textwidth}
        \includegraphics[height=1.3in]{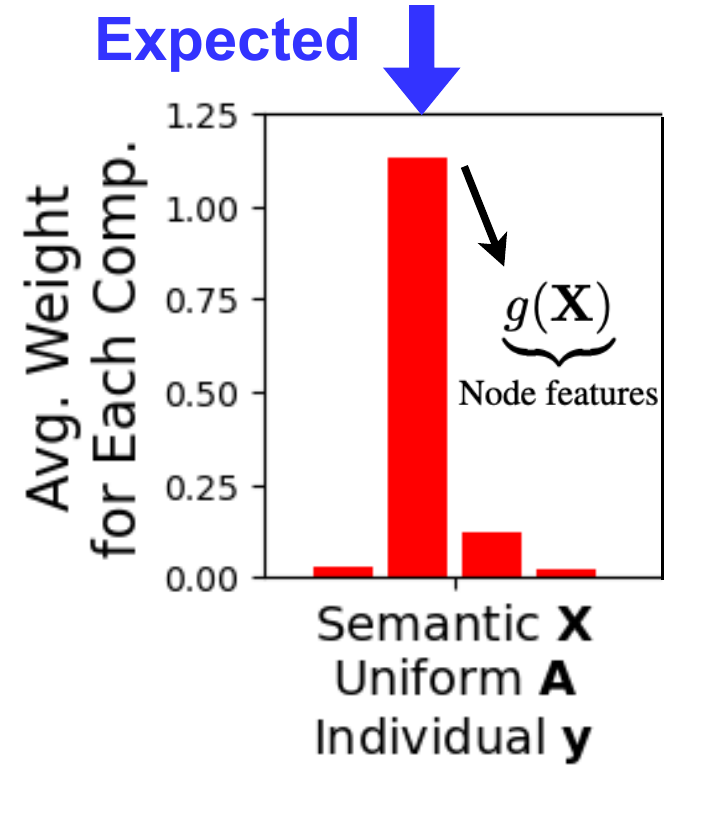}
        \spaceAboveSubfigureCaption
        \caption{No network effects}
        \label{fig:interpret-1}
    \end{subfigure}
    \begin{subfigure}{0.24\textwidth}
        \includegraphics[height=1.3in]{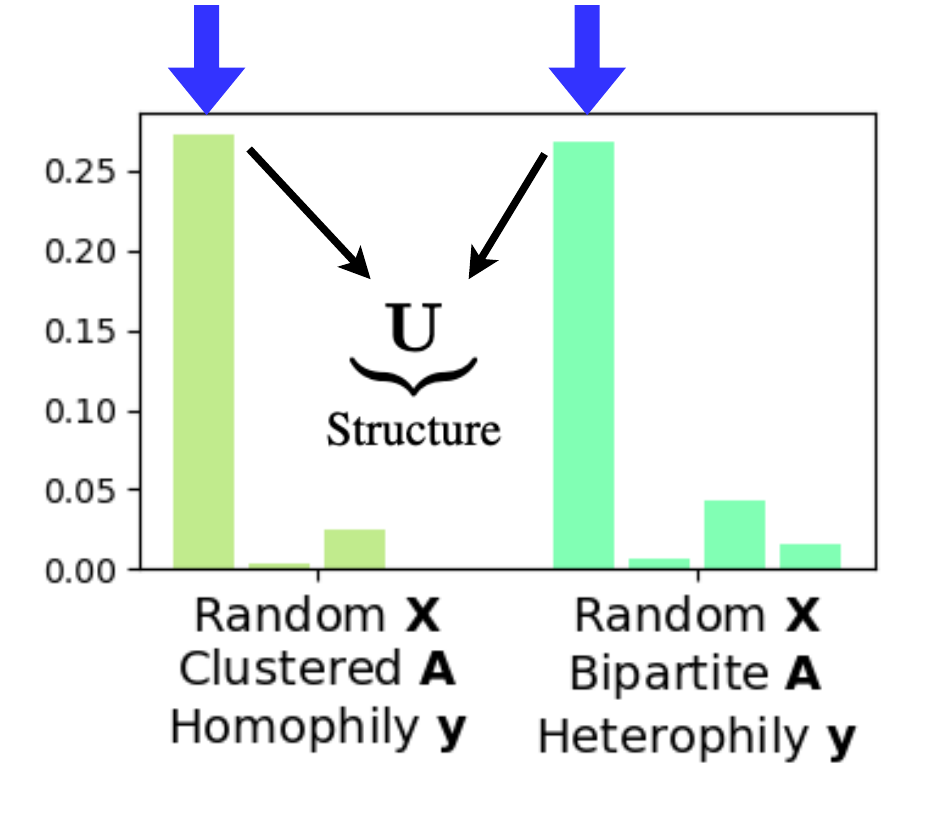}
        \spaceAboveSubfigureCaption
        \caption{Useless features}
        \label{fig:interpret-2}
    \end{subfigure}
    
    \spaceabovefigurecaption
    \caption{
        \myTag{\method is interpretable}: it suppresses useless information and focuses on informative ones for each scenario: (a) self-features $g(\mathbf{X})$ and (b) structural features $\mathbf{U}$.
        The learned weights are directly matched with the expectations.
    }
    \label{fig:interpret}
\spaceBelowSmallFigure
\end{figure}


\subsection{Interpretability (RQ4)}

Figure~\ref{fig:interpret} illustrates the weights learned by the classifier in \method for the sanity checks, where the ground truths are known.
\method assigns large weights to the correct factors in graphs with different mutual information between variables.
When there are no network effects in Figure~\ref{fig:interpret-1}, it successfully assigns the largest weights to the self-features $g(\mathbf{X})$, ignoring all other components.
When the features are useless in Figure~\ref{fig:interpret-2}, it puts most of the attention on the structural features $\mathbf{U}$, which does not rely on $\mathbf{X}$.

\subsection{Ablation Studies (RQ5)}

We run two types of ablation studies to support the main ideas of \method: its design decisions and two-hop aggregation.

\header{Design decisions}
In Table \ref{table:ablnl}, we show the accuracy of \method when each of its core ideas is disabled: sparse regularization, PCA, and the structural features $\mathbf{U}$.
\method performs best with all of its ideas enabled; it is always included in the top two in each dataset.
This shows that \method is designed effectively with ideas that help improve its accuracy on real-world datasets.
Note that the sparse regularization and PCA improve the efficiency of \method, by reducing the number of parameters, as well as its accuracy.

\header{Receptive fields}
To analyze the effect of changing the receptive field of \method, we vary the distance of aggregation (i.e., the value of $K$) in Table~\ref{table:ablrf}.
The values of $k_{\text{row}}$ or $k_{\text{sym}}$ given as sets, e.g., $\{2, 4, 6\}$, represent that we include more than one component to \method with different $K$, increasing the overall complexity of decisions.
Since $\mathbf{A}_\mathrm{row}$ is designed to consider heterophily relations, we use only the even values of $k_{\text{row}}$.
Table~\ref{table:ablrf} shows no significant gain in accuracy by increasing the values of $k_{\text{row}}$ and $k_{\text{sym}}$, or even including more components to \method; the accuracies of all variants are similar to each other in all cases.
That is, \method works sufficiently well even with the $2$-step aggregation for both $\mathbf{A}_\mathrm{row}$ and \smash{$\tilde{\mathbf{A}}_\mathrm{sym}$}.

%% file: 070conclusion.tex
\section{Conclusion}
\label{sec:conclusion}

We propose \method, a simple but effective model for semi-supervised node classification, which is designed by the \principleLong principle.
We summarize our contributions as follows:
\begin{compactitem}
    \item \mycontribution{C1 - Method:}
        \method outperforms state-of-the-art GNNs in both synthetic and real datasets, showing the best robustness; \method succeeds in all types of graphs with homophily, heterophily, noisy features, etc.
        \method is scalable to million-scale graphs, even when other baselines run out of memory, making interpretable decisions from its linearity.
    \item \mycontribution{C2 - Explanation:}
        Our \framework framework illuminates the fundamental similarities and differences of popular GNN variants (see Table~\ref{table:lin-gnn}), revealing their pain points.
    \item \mycontribution{C3 - Sanity checks:}
        Our sanity checks immediately highlight the strengths and weaknesses of each GNN method before it is sent to production (see Table~\ref{table:sanity-tests}).
    \item \mycontribution{C4 - Experiments:}
        Our extensive experiments explain the success of \method in various aspects: linearity, robustness, receptive fields, and ablation studies (see Tables \ref{table:expr} to \ref{table:ablrf}). 
\end{compactitem}

\header{Reproducibility}
Our code, along with our datasets for \emph{sanity checks}, is available at \codeurl.

%% file: 110proof.tex
\section{Proofs of Lemmas}

\subsection{Proof of Lemma \ref{lemma:linear-models}}
\label{appendix:proof-linear-models}

\begin{proof}
We prove the lemma for each of SGC, DGC, S$^2$GC, and G$^2$CN.
The propagator function of SGC \citep{Wu19SGC} directly fits the definition of linearization.
DGC \citep{Wang21DGC} has variants DGC-Euler and DGC-DK.
We focus on DGC-Euler, which is mainly used in their experiments.
Then, DGC also fits the definition of linearization.
S$^2$GC \citep{Zhu21S2GC} computes the summation of features propagated with different numbers of steps.
The original formulation divides the added features by $K$, which is safely ignored as we multiply the weight matrix $\mathbf{W}$ to the transformed feature for classification.

G$^2$CN \citep{Li22G2CN} does not provide an explicit formulation of the propagator function.
The parameterized version $\mathcal{P}'$ of the propagator function is $\mathcal{P}'(\mathbf{A}, \mathbf{X}; \{ \theta_i \}_{i=1}^N) =
        \sum_{i=1}^N \theta_i \mathbf{H}_{i, K}$,
where $\theta_i$ is a parameter.
The $k$-th feature representation $\mathbf{H}_{i, k}$ is recursively defined as
\smash{$\mathbf{H}_{i, k} = [\mathbf{I} - \frac{T_i}{K} ((b_i - 1) \mathbf{I} + \mathbf{A}_\mathrm{sym})^2] \mathbf{H}_{i, k-1}$}, where $\mathbf{L} = \mathbf{I} - \mathbf{A}_\mathrm{sym}$ is the normalized Laplacian matrix, $N$, $T_i$, and $b_i$ are hyperparameters, and $\mathbf{H}_{i, 0} = \mathbf{X}$.
Then, we make it contain no learnable parameters as $\mathcal{P}(\mathbf{A}, \mathbf{X}) = \concat_{i=1}^N \mathbf{H}_{i, K}$.
%
We prove the lemma from the four cases.
\end{proof}

\subsection{Proof of Lemma \ref{lemma:complexity}}
\label{appendix:proof-complexity}

\begin{proof}
    The training of \method consists of three parts: SVD, PCA, and LR.
    The time complexity of SVD is $O(de + d^2n)$ as \method runs the sparse truncated SVD for generating structural features.
    The complexity of PCA, which is applied to each of the four components in \Eqref{eq:proposed}, is $O(d^2n + d^3)$.
    The complexity of the gradient-based optimization of LR is $O(dnt)$, where $t$ is the number of epochs.
    We safely assume $t < n$, since $t < 100$ in all our experiments.
    We prove the lemma by combining the three complexity terms.
\end{proof}

%% file: 120linear.tex
\section{Linearization Processes}
\label{appendix:linearization}

\subsection{Linearization of Decoupled GNNs}
\label{appendix:linearization-decoupled}



\begin{lemma}
PPNP \citep{Klicpera19APPNP} is linearized by \framework as
\begin{equation}
    \mathcal{P}(\mathbf{A}, \mathbf{X}) = (\mathbf{I}_n - (1 - \alpha) \AsymK{})^{-1} \mathbf{X},
\end{equation}
where $0 < \alpha < 1$ controls the weight of self-loops.
\end{lemma}

\begin{proof}
\spaceAroundProof
The proof is straightforward, since the authors present a closed-form representation of the propagator function.
We remove the activation function from the original representation.
\spaceAroundProof
\end{proof}


\begin{lemma}
APPNP \citep{Klicpera19APPNP} is linearized by \framework as
\begin{equation}
    \mathcal{P}(\mathbf{A}, \mathbf{X}) = \left[ {\textstyle \sum_{k=0}^{K-1}} \alpha (1 - \alpha)^k \AsymK{k} + (1 - \alpha)^K \AsymK{K} \right] \mathbf{X},
\label{eq:appnp}
\end{equation}
where $0 < \alpha < 1$ controls the weight of self-loops.
\end{lemma}

\begin{proof}
\spaceAroundProof
We assume that the initial node representation is created by a single linear layer, i.e., $\mathbf{X} \mathbf{W}$.
Then, the $k$-th representation matrix $\mathbf{H}_k$ is represented as $\mathbf{H}_k =
        (1 - \alpha) \AsymK{} \mathbf{H}_{k-1} + \alpha \mathbf{X} \mathbf{W}$,
where $0 < \alpha < 1$ is a hyperparameter.
We derive the closed-form representation of $\mathbf{H}_K$ and remove redundant parameters.
\spaceAroundProof
\end{proof}


\begin{lemma}
GDC \citep{Klicpera19GDC} is linearized by \framework as
\begin{equation}
    \mathcal{P}(\mathbf{A}, \mathbf{X}) = \tilde{\mathbf{S}}_\mathrm{sym}' \odot \mathbf{1}(\tilde{\mathbf{S}}_\mathrm{sym}' \geq \epsilon),
\label{eq:gdc}
\end{equation}
where $\mathbf{S} = \sum_{k=0}^\infty \alpha (1 - \alpha)^k \AsymK{k}$, $\odot$ is the elementwise multiplication, $\mathbf{1}$ is a matrix that contains one if each element holds the condition and zero if otherwise, and $\alpha$ and $\epsilon$ are hyperparameters.
\end{lemma}

\begin{proof}
\spaceAroundProof
GDC presents various forms of propagation functions by generalizing APPNP.
We pick the most representative one given in the paper, which is directly related to APPNP.
The unnormalized version of the propagation matrix is $\mathbf{S}' = \sum_{k=0}^\infty \alpha (1 - \alpha)^k \AsymK{k}$,
which is then normalized and sparsified as \smash{$\mathbf{S} = \mathrm{sparsify}(\tilde{\mathbf{S}}_\mathrm{sym}')$}.
The matrix $\tilde{\mathbf{S}}_\mathrm{sym}'$ represents adding self-loops and applying the symmetric normalization to $\mathbf{S}'$.
The paper gives two approaches for sparsification, which are a) removing elements smaller than $\epsilon$, which is a hyperparameter, and b) selecting the top $k$ neighbors for each node.
We take the first approach, which is easier to represent in a closed form, getting \Eqref{eq:gdc}.
\spaceAroundProof
\end{proof}


\begin{lemma}
GPR-GNN \citep{Chien21GPRGNN} is linearized by \framework as
\begin{equation}
    \mathcal{P}(\mathbf{A}, \mathbf{X}) =
        {\textstyle \concat_{k=0}^K} \AsymK{k} \mathbf{X}.
\label{eq:gpr-gnn}
\end{equation}
\end{lemma}

\begin{proof}
\spaceAroundProof
We assume that the initial node representation is created by a single linear layer, i.e., $\mathbf{X} \mathbf{W}$.
Then, we replace the summation in the original propagator function with concatenation to remove the learnable parameters, getting \Eqref{eq:gpr-gnn}.
\spaceAroundProof
\end{proof}

\subsection{Linearization of Coupled GNNs}
\label{appendix:linearization-coupled}



\begin{lemma}
ChebNet \citep{Defferrard16ChebNet} is linearized by \framework as
\begin{equation}
    \mathcal{P}(\mathbf{A}, \mathbf{X}) = {\textstyle \concat_{k=0}^{K-1}} \mathbf{A}_\mathrm{sym}^k \mathbf{X}.
\label{eq:chebnet}
\end{equation}
\end{lemma}

\begin{proof}
Let $\mathbf{H}_k$ be the $k$-th node representation matrix, and $\mathbf{L} = \mathbf{I} - \mathbf{A}_\mathrm{sym}$ be the graph Laplacian matrix normalized symmetrically.
Then, the propagator function of ChebNet with parameters $\theta$ is $\mathcal{P}'(\mathbf{A}, \mathbf{X}; \theta) = \sum_{k=0}^{K-1} \theta_k \mathbf{H}_k$
%
where $\mathbf{H}_k = a_k \mathbf{A}_\mathrm{sym}^k \mathbf{X} + a_{k-1} \mathbf{A}_\mathrm{sym}^{k-1} \mathbf{X} + \cdots + a_0 \mathbf{X}$, and $a_0, \cdots, a_k$ are constants.
Since we have $K$ free parameters $\theta_0, \cdots, \theta_K$, we safely rewrite the propagator function as $\mathcal{P}'(\mathbf{A}, \mathbf{X}; \theta) = \sum_{k=0}^{K-1} \theta_k \mathbf{A}_\mathrm{sym}^k \mathbf{X}$.
We remove the parameters by replacing the summation with concatenation.
\end{proof}


\begin{lemma}
GraphSAGE \citep{Hamilton17SAGE} is linearized by \framework as
\begin{equation}
    \mathcal{P}(\mathbf{A}, \mathbf{X}) = {\textstyle \concat_{k=0}^K} \mathbf{A}_\mathrm{row}^k \mathbf{X}.
\end{equation}
\end{lemma}

\begin{proof}
\spaceAroundProof
We assume the mean aggregator of GraphSAGE among various choices.
By removing the activation function, each layer of GraphSAGE is linearized as $\mathcal{F}(\mathbf{X}) = \mathbf{X} \mathbf{W}_1 + \mathbf{A}_\mathrm{row} \mathbf{X} \mathbf{W}_2$,
where $\mathbf{A}_\mathrm{row} = \mathbf{D}^{-1} \mathbf{A}$ represents the mean operator in the aggregation, and $\mathbf{W}_1$ and $\mathbf{W}_2$ are learnable weight matrices.
If we stack $K$ layers with reparametrization, we get \smash{$\mathcal{F}^K(\mathbf{X}) = \sum_{k=1}^K \mathbf{A}_\mathrm{row}^k \mathbf{X} \mathbf{W}_k$}.
Note that a different weight matrix $\mathbf{W}_k$ is applied to each layer $k$.
This is equivalent to concatenating the transformed features of all layers and learning a single large weight matrix in training.
\spaceAroundProof
\end{proof}


\begin{lemma}
GCNII \citep{Chen20GCNII} is linearized by \framework as
\begin{equation}
    \mathcal{P}(\mathbf{A}, \mathbf{X}) =
        {\textstyle \concat_{k=0}^{K-2}} \AsymK{k} \mathbf{X} \cat ((1 - \alpha) \AsymK{K} + \alpha \AsymK{K-1}) \mathbf{X},
\label{eq:gcnii}
\end{equation}
where $\alpha$ is a hyperparameter.
\end{lemma}

\begin{proof}
\spaceAroundProof
After removing the activation function, the $l$-th layer $\mathcal{F}_l$ of GCNII is $\mathcal{F}_l(\mathbf{H}) = ((1 - \alpha_l) \AsymK{} \mathbf{H} + \alpha_l \mathbf{X})) ((1 - \beta_l) \mathbf{I} + \beta_l \mathbf{W}_l)$,
where $\alpha_l$ and $\beta_l$ are hyperparameters, and $\mathbf{W}_l$ is a weight matrix.
The second term is equivalent to $\mathbf{W}_l$ regardless of the value of $\beta_l$, since $\mathbf{W}_l$ is a free parameter.
We also set $\alpha_l$ to a constant $\alpha$ which is the same for every layer $l$, following the original paper \citep{Chen20GCNII}.\footnote{$\alpha$ is set to $0.1$ in the original paper of GCNII \citep{Chen20GCNII}.}
Then, the equation is simplified as $\mathcal{F}_l(\mathbf{H}) = ((1 - \alpha) \AsymK{} \mathbf{H} + \alpha \mathbf{X}) \mathbf{W}_l$.
The closed-form representation is $\mathcal{F}_K(\mathbf{X}) =
        (1 - \alpha)^{K-1} ((1 - \alpha) \AsymK{K} + \alpha \AsymK{K-1}) \mathbf{X} \mathbf{W}_K +
        \alpha \sum_{k=0}^{K-2} (1 - \alpha)^k \AsymK{k} \mathbf{X} \mathbf{W}_k$.
We safely remove the constants that can be included in the weight matrices, and replace the summation with concatenation.
%
\spaceAroundProof
\end{proof}


\begin{lemma}
H$_2$GCN \citep{Zhu20H2GCN} is linearized by \framework as
\begin{equation}
    \mathcal{P}(\mathbf{A}, \mathbf{X}) = {\textstyle \concat_{k=0}^{2K}} \mathbf{A}_\mathrm{sym}^k \mathbf{X}.
\end{equation}
\end{lemma}

\begin{proof}
\spaceAroundProof
H$_2$GCN concatenates the features generated from every layer, each of which also concatenates the features averaged for the one- and two-hop neighbors from the previous layer.
The self-loops are not added to $\mathbf{A}$ during the propagations.
\spaceAroundProof
\end{proof}

\subsection{Partial Linearization of Attention GNNs}
\label{appendix:linearization-attention}



\begin{lemma}
DA-GNN \citep{Liu20DAGNN} is partially linearized by \framework as
\begin{equation}
    \mathcal{P}(\mathbf{A}, \mathbf{X}) = {\textstyle \sum_{k=0}^K} \mathrm{diag}(\AsymK{k} \mathbf{X} \mathbf{w}) \AsymK{k} \mathbf{X}.
\label{eq:dagnn}
\end{equation}
where $\mathrm{diag}(\cdot)$ is a function that generates a diagonal matrix from a vector, and $\mathbf{w}$ is a learnable parameter.
\end{lemma}

\begin{proof}
\spaceAroundProof
DA-GNN has separate feature transformation and propagation steps as in the decoupled models.
We assume that the initial node representation is created by a single linear layer, i.e., $\mathbf{X} \mathbf{W}$.
Then, the $k$-th representation matrix is $\mathbf{H}_k = \AsymK{k} \mathbf{X} \mathbf{W}$.
DA-GNN computes the weighted sum of representations for all $k \in [0, K]$, where the weight values are determined also from the representation matrices: $\mathcal{P}(\mathbf{A}, \mathbf{X}) = \sum_{k=0}^K \mathrm{diag}(\mathbf{H}_k \mathbf{s}) \mathbf{H}_k$,
where $\mathbf{s}$ is a learnable weight vector.
Lastly, we remove the redundant parameters.
\spaceAroundProof
\end{proof}


\begin{lemma}
GAT \citep{Velickovic18GAT} is partially linearized by \framework as
\begin{equation}
    \mathcal{P}(\mathbf{A}, \mathbf{X}) = {\textstyle \prod_{k=1}^K} \left[ \mathrm{diag}(\mathbf{X} \mathbf{w}_{k,1}) \tilde{\mathbf{A}} + \tilde{\mathbf{A}} \mathrm{diag}(\mathbf{X} \mathbf{w}_{k,2}) \right] \mathbf{X},
\label{eq:gat}
\end{equation}
where $\mathbf{w}_{k, 1}$ and $\mathbf{w}_{k, 2}$ are learnable weight vectors.
\end{lemma}

\begin{proof}
\spaceAroundProof
We apply the following changes to linearize GAT, whose linearization is not straightforward due to the nonlinearity in the attention function:
(a) We simplify the attention function, removing the exponential and normalization: $\alpha_{ij} = \exp(e_{ij}) / \sum_k \exp(e_{ik}) \approx e_{ij}.$
(b) We remove LeaklyReLU in the computation of $e_{ij}$.
(c) We assume the single-head attention.
Then, the edge weight $e_{ij}$, which is the $(i, j)$-th element of the propagator matrix, is defined as $e_{ij} = \mathbf{a}_\mathrm{dst}^\top (\mathbf{W}^\top \mathbf{x}_i) + \mathbf{a}_\mathrm{src}^\top (\mathbf{W}^\top \mathbf{x}_j)$,
where $\mathbf{x}_i$ and $\mathbf{x}_j$ are feature vectors of length $d$ for node $i$ and $j$, respectively, $\mathbf{W}$ is a $d \times c$ weight matrix, and $\mathbf{a}_\mathrm{dst}$ and $\mathbf{a}_\mathrm{src}$ are learnable weight vectors of length $c$.

We derive the initial form of a linearized GAT layer as $\mathbf{H} =
        [ \mathrm{diag}(\mathbf{X} \mathbf{W} \mathbf{a}_\mathrm{dst}) \tilde{\mathbf{A}} + \tilde{\mathbf{A}} \mathrm{diag}(\mathbf{X} \mathbf{W} \mathbf{a}_\mathrm{src}) ] \mathbf{X} \mathbf{W}$.
%
Since all $\mathbf{a}_\mathrm{dst}$, $\mathbf{a}_\mathrm{src}$, and $\mathbf{W}$ are free parameters, we generalize it as $\mathbf{H} =
        [ \mathrm{diag}(\mathbf{X} \mathbf{w}_\mathrm{dst}) \tilde{\mathbf{A}} + \tilde{\mathbf{A}} \mathrm{diag}(\mathbf{X} \mathbf{w}_\mathrm{src}) ] \mathbf{X} \mathbf{W}$,
where $\mathbf{w}_\mathrm{dst}$ and $\mathbf{w}_\mathrm{src}$ are learnable vectors that replace $\mathbf{a}_\mathrm{dst}$ and $\mathbf{a}_\mathrm{src}$, respectively.
Then, we get \Eqref{eq:gat}.
\spaceAroundProof
\end{proof}


%% file: 130sanity.tex
\section{Implementation of Sanity Checks}
\label{appendix:sanity-checks}

There exist various ways to generate synthetic graphs satisfying the requirements of our sanity checks \citep{Leskovec10Kronecker, Barabasi99Science}.
However, we choose the simplest approach to focus on the mutual information between variables, rather than other characteristics of real-world graphs.

\header{Structure}
We assume that the number of node clusters is the same as the number $c$ of labels.
We divide all nodes into $c$ groups and then decide the edge densities for intra- and inter-connections of groups based on the type: uniform, homophily, and heterophily.
The expected number of edges is the same for all three cases.

A notable characteristic of a heterophily structure $\mathbf{A}$ is that $\mathbf{A}^2$ follows homophily.
If we create inter-group connections for all pairs of different groups, it creates noisy $\mathbf{A}^2$ with inter-group connections.
For consistency, we set the number of classes to an even number in our experiments, randomly pick paired classes such as $(1, 3)$ and $(2, 4)$ when $c=4$, for example, and create inter-group connections only for the chosen pairs.
In this way, we create a non-diagonal block-permutation matrix $\mathbf{A}$, as shown in Figure \ref{fig:adj-heterophily}.

\header{Features}
We assume that every feature element $x_{ij}$ is basically sampled from a uniform distribution.
In the random case, we sample each element from the distribution $\mathcal{U}(0, 1)$ between 0 and 1.
In the structural case, we run the low-rank support vector decomposition (SVD) \citep{Halko11SVD} to make $\mathbf{X}$ have structural information.
Given $\mathbf{U}\Sigma\mathbf{V}^\top \approx \mathbf{A}$ from the low-rank SVD, we take $\mathbf{U}$ and normalize each column to have the zero-mean and unit-variance.
The rank $r$ in the SVD is a hyperparameter; higher $r$ captures the structure better but can give noisy information.
We also apply ReLU to make $\mathbf{U}$ positive.

In the semantic case, we randomly pick $c$ representative vectors $\{\mathbf{v}_k\}_{k=1}^c$ from the uniform distribution, which correspond to the $c$ different classes.
Then, for each node $i$ with label $y$, we sample a feature vector $\mathbf{x}_i$ such that $\argmax_k \mathbf{x}_i^\top \mathbf{v}_k = y$.
In this way, we have random vectors that have sufficient semantic information for the classification of labels, with a guarantee that the perfect linear decision boundaries can be drawn in the feature space $\mathbf{X}$.

%% file: 140exp.tex
\section{Hyperparameters}
\label{appendix:reproducibility}

Table \ref{tab:hyper} summarizes the search space of hyperparameters for \method and the competitors.
We conduct a grid search based on the validation accuracy for each data split for a fair comparison between different models.
It is notable that we run all our experiments on five different random seeds, and thus the optimal set of hyperparameters can be found differently for those five runs.


\begin{table}
    \centering
    \caption{
        Search space of hyperparameters.
    }
    \resizebox{0.47\textwidth}{!}{
    \begin{tabular}{l|l}
      \toprule
      \textbf{Method} & \textbf{Hyperparameters} \\
      \midrule
      LR & $wd=[0, 5e^{-4}]$ \\
      \midrule
      SGC & $wd=[0, 5e^{-4}], K=2$ \\
      DGC & $wd=[0, 5e^{-4}], K=200, T=[3, 4, 5, 6]$ \\
      S$^2$GC & $wd=[0, 5e^{-4}], K=16, \alpha=[0.01, 0.03, 0.05, 0.07, 0.09]$ \\
      G$^2$CN & $wd=[0, 5e^{-4}], K=100, N=2, T_{1}=T_{2}=[10, 20, 30, 40], b_{1}=0, b_{2}=2$ \\
      \midrule
      GCN & $wd=[0, 5e^{-4}], lr=[2e^{-3}, 0.01, 0.05], K=2$ \\
      SAGE & $wd=[0, 5e^{-4}], lr=[2e^{-3}, 0.01, 0.05], K=2$ \\
      GCNII & $wd=[0, 5e^{-4}], lr=0.01, K=[8, 16, 32, 64], \alpha=[0.1, 0.2, 0.5], \theta=[0.5, 1, 1.5]$ \\
      H$_2$GCN & $wd=[0, 5e^{-4}], lr=[2e^{-3}, 0.01, 0.05], K=[1, 2]$ \\
      APPNP & $wd=[0, 5e^{-4}], lr=[2e^{-3}, 0.01, 0.05], K=10, \alpha=0.1$ \\
      GPR-GNN & $wd=[0, 5e^{-4}], lr=[2e^{-3}, 0.01, 0.05], K=10, \alpha=[0.1, 0.2, 0.5, 0.9]$ \\
      GAT & $wd=[0, 5e^{-4}], lr=[2e^{-3}, 0.01, 0.05], K=2, heads=8$ \\
      \midrule
      \method & $wd_{1}=[1e^{-3}, 1e^{-4}, 1e^{-5}], wd_{2}=[1e^{-3}, 1e^{-4}, 1e^{-5}, 1e^{-6}]$ \\
      \bottomrule
    \end{tabular}
    \label{tab:hyper}
    }
\end{table}
